\documentclass [sigconf]{acmart}
\setcopyright{none}
\renewcommand\footnotetextcopyrightpermission[1]{}
\acmConference[WSDM '27]
{The 20th ACM International Conference on Web Search and Data Mining}
{February 15--19, 2027}
{Hong Kong SAR, China}
\AtBeginDocument{%
  }

\usepackage{amsmath}
\usepackage{caption}
\usepackage{subcaption}
\usepackage{pifont}
\usepackage{multicol}
\usepackage{multirow}
\usepackage{tabu}
\usepackage{todonotes}
\usepackage{booktabs}
\usepackage{mathtools}
\usepackage{adjustbox}
\usepackage{enumitem}
\usepackage[table]{xcolor}
\usepackage{colortbl}
\usepackage{ragged2e}
\usepackage{graphicx}
\usepackage{subcaption}
\usepackage{enumitem}
\usepackage{bbm}

\newcommand{\para}[1]{{\vspace{3.5pt} \bf \noindent #1 \hspace{8pt}}}

\begin{document}

% \title{Bridging Supervised and Self-Supervised Representations for Out-of-Distribution Graph Learning}
\title{Complementary Supervised and Self-Supervised Representations for Out-of-Distribution Graph Learning}

\author{Qingying Hao}
\affiliation{%
  \institution{ShanghaiTech University}
  \city{Shanghai}
  \country{China}
}
\email{haoqy@shanghaitech.edu.cn}

\author{Zikang Chen}
\affiliation{%
  \institution{ShanghaiTech University}
  \city{Shanghai}
  \country{China}
}
\email{chenzk2025@shanghaitech.edu.cn}

\author{Chuxuan Hu}
\affiliation{%
  \institution{University of Illinois Urbana-Champaign}
  \city{Urbana}
  \state{Illinois}
  \country{USA}
}
\email{chuxuan3@illinois.edu}

\author{Jinyuan Jia}
\affiliation{%
  \institution{The Pennsylvania State University}
  \city{University Park}
  \state{Pennsylvania}
  \country{USA}
}
\email{jinyuan@psu.edu}

\author{Bo Li}
\affiliation{%
  \institution{University of Illinois Urbana-Champaign}
  \city{Urbana}
  \state{Illinois}
  \country{USA}
}
\email{lbo@illinois.edu}

\author{Gang Wang}
\affiliation{%
  \institution{University of Illinois Urbana-Champaign}
  \city{Urbana}
  \state{Illinois}
  \country{USA}
}
\email{gangw@illinois.edu}

\author{Carl Gunter}
\affiliation{%
  \institution{University of Illinois Urbana-Champaign}
  \city{Urbana}
  \state{Illinois}
  \country{USA}
}
\email{cgunter@illinois.edu}
\begin{abstract}

Out-of-distribution (OOD) generalization remains challenging for graph
neural networks (GNNs), as graph distributions can vary substantially
across time and domains. Supervised and self-supervised graph
representation learning are guided by distinct objectives and offer
different perspectives on graph representations. In this work, we study
whether self-supervised representations (SSL) can provide complementary
signals to improve supervised OOD node classification. We develop two
backbone-agnostic frameworks that exploit such information at different
stages of learning and prediction. Co-Train jointly learns
supervised and SSL representations and adaptively integrates them during
training, while Dual-Space Retrieval performs non-parametric prediction in the two representation spaces and combines their predictions through confidence-aware fusion at inference time. The supervised and SSL encoders are separately parameterized and
need not share the same GNN architecture.

We evaluate multiple GNN backbones and two distinct SSL objectives,
DGI and GRACE, on four graph benchmarks spanning temporal and
cross-domain distribution shifts. Extensive experiments show that
Co-Train consistently outperforms strong supervised OOD baselines,
while Dual-Space Retrieval achieves competitive performance as a
flexible non-parametric alternative. Results across different backbones
and SSL objectives, together with representation analyses and ablations,
demonstrate that SSL representations provide complementary information
to supervised representations and can improve OOD node classification
across diverse settings.

\end{abstract}



\maketitle

\section{Introduction}
\label{sec:intro}

Graph-structured data are ubiquitous in real-world applications,
including social platforms, financial systems, traffic networks, and
molecular analysis. Graph Neural Networks (GNNs) have become a
fundamental approach for learning from such data by capturing relational
patterns and structural dependencies. They have been widely applied to
large-scale recommendation \cite{ying2018pinsage}, social network
analysis \cite{goel2019end}, and molecular property prediction
\cite{buterez2024transfer}.

Most GNNs, however, assume that training and test data follow the same
distribution, an assumption that rarely holds in practice
\cite{mario_unsupervised,zhu2021shift}. Graph distributions may evolve
over time or vary substantially across domains. For example, transaction
patterns in financial networks can change over time, while social
networks from different communities may exhibit distinct structural and
feature distributions. Such shifts can substantially degrade GNN
performance \cite{real-world-problem1,zhang2022dynamic}, motivating
graph out-of-distribution (OOD) generalization.

Existing graph OOD methods have largely focused on supervised learning.
A common strategy constructs multiple training environments through
graph augmentations and encourages models to learn representations that
remain predictive across these environments
\cite{wu2022dir,wu2022eerm,lisa}. Other approaches use causal analysis
to separate label-relevant factors from spurious correlations
\cite{chen2022ciga,wu2024canet}. While effective, supervised
representations are optimized directly toward training labels and may
capture label-correlated patterns that do not remain stable under
distribution shifts
\cite{zhu2021shift,fanGraphCorrelation,liu2022towards}.

Self-supervised learning (SSL) offers a complementary source of
representation information. Methods such as DGI~\cite{dgi} and
GRACE~\cite{zhu2020grace} learn graph representations without label
supervision by exploiting structural context or consistency across
augmented graph views. However, SSL representations are not explicitly
optimized for the downstream classification task. Graph augmentations
may discard task-relevant information~\cite{trivedi2021augmentations},
while SSL representations may retain factors unrelated to the target
labels~\cite{shen2024beyond,liu2022towards}. These different learning
objectives motivate us to investigate whether SSL representations can
provide useful complementary information to supervised representations
for graph OOD generalization.

Our empirical analyses support this complementarity: supervised and SSL
representations provide distinct signals, and integrating them improves
OOD generalization across multiple datasets and settings. Based on this
observation, we develop two mechanisms that incorporate SSL information
at different stages of prediction: \textit{training-time adaptive
representation integration} and \textit{inference-time confidence-aware
dual-space retrieval}.

Our first framework, \textbf{Co-Train}, exploits this complementarity
during training. It uses separately parameterized supervised and SSL
encoders and jointly optimizes their representations for downstream
classification. A gated fusion mechanism adaptively incorporates the SSL
representation into the supervised representation, while an alignment
objective encourages correspondence between the two representation
spaces. Our second framework, \textbf{Dual-Space Retrieval}, operates at
inference time without joint optimization of the two branches. It
performs non-parametric prediction independently in the supervised and
SSL spaces and combines their similarity-weighted prediction scores
through confidence-aware fusion. Depending on the prediction setting,
we use neighbor-based or class-centroid retrieval to obtain label
signals from each representation space.

Both frameworks are backbone-agnostic: the supervised and SSL encoders
do not share parameters and are not required to use the same GNN
architecture. We instantiate the SSL branch with DGI and GRACE to
evaluate whether the benefits generalize across different SSL
objectives, and further evaluate the frameworks with multiple GNN
backbones. This modular design allows SSL representations to complement
supervised OOD learning without constraining either branch to a
particular backbone or SSL objective.

We evaluate our frameworks on four graph OOD benchmarks,
\texttt{OGB-Arxiv}, \texttt{Elliptic}, \texttt{Twitch-Explicit}, and
\texttt{Facebook-100}, covering both temporal and cross-domain
distribution shifts. Co-Train consistently improves over strong
supervised OOD baselines across datasets, backbone architectures, and
SSL objectives. Dual-Space Retrieval also achieves competitive
performance across diverse settings, often providing the second-best
results while offering a non-parametric alternative to joint training.
Representation analyses and extensive ablations further demonstrate the
benefit of combining supervised and SSL representations and characterize
when each design is most effective. Co-Train is computationally
efficient, while the retrieval cost of Dual-Space Retrieval depends on
the retrieval set and can be substantially reduced through class-centroid
retrieval.

\para{Contributions.}
Our main contributions are:
\begin{itemize}[leftmargin=*, noitemsep, topsep=0pt]

\item We empirically establish the complementarity of supervised and
self-supervised representations for graph OOD generalization, showing
that SSL representations can provide useful information beyond
label-guided supervised representations.

\item Based on this observation, we develop two backbone-agnostic
frameworks that exploit this complementarity at different stages:
\textbf{Co-Train}, which adaptively integrates supervised and SSL
representations during training, and \textbf{Dual-Space Retrieval},
which performs confidence-aware non-parametric prediction over the two
representation spaces at inference time.

\item Extensive experiments across temporal and cross-domain shifts,
multiple GNN backbones, and two SSL objectives demonstrate consistent
improvements over strong supervised OOD baselines. Representation
analyses, ablations, and sensitivity studies further validate the
benefits of combining the two representation spaces and the
generalizability of our designs across different model configurations.

\item Co-Train provides an efficient training-time integration mechanism,
while Dual-Space Retrieval offers a flexible non-parametric alternative
whose retrieval cost can be substantially reduced through class-centroid
representations.
\end{itemize}
\section{Related Work}
\label{sec:related}

\subsection{Graph Neural Networks}
Graph neural networks (GNNs) have been widely used for graph-structured data. One class of methods builds on spectral graph theory, where graph convolution is defined in the frequency domain \cite{gcn, defferrard2016convolutional}. For example, GCN \cite{gcn} performs convolutional operation using a first-order Chebyshev polynomial approximation. Another class of methods is spatial-based GNNs, which define graph convolution over node neighborhoods. GraphSAGE \cite{graphsage} learns representation by sampling and aggregating information from local neighborhoods, while GAT \cite{gat} incorporates attention mechanisms to learn weights for neighbors.
There are other works that solve vanishing gradients and gradient smoothing in deep GNNS, GCNII \cite{gcn2} introduces residual connections and identity mapping, while GPR-GNN \cite{gpr} earns a flexible propagation scheme by weighting multi-hop neighborhood information, generalizing Personalized PageRank with learnable coefficients. Additional efforts have focused on other aspects, such as enhancing scalability \cite{graphsaint} and improving the trustworthiness of GNNs. 
\subsection{Distribution Shift Detection in Graph Neural Networks}

\para{Supervised Learning Approaches.}
A range of supervised methods have been proposed for graph
out-of-distribution (OOD) generalization, often by learning invariant
representations across distribution shifts. ERM~\cite{erm} minimizes
classification loss over the training data without explicitly accounting
for distribution shifts. DIR~\cite{wu2022dir} constructs synthetic
environments by sampling diverse subgraphs and applies causal analysis
to learn invariant predictors, while EERM~\cite{wu2022eerm} dynamically
generates environments through reinforcement learning and minimizes
prediction risks across them. LiSA~\cite{lisa} generates label-invariant subgraphs and learns a shift-invariant encoder through an energy-based objective. GRM~\cite{wang2025generative} further models invariant information by generating continuous invariant subgraphs. Other supervised approaches introduce additional mechanisms:
GNNSafe~\cite{wu2023gnnsafe} calibrates prediction confidence using
energy functions, while KnowGraph~\cite{andyZhouKG} incorporates
symbolic reasoning and domain knowledge for graph anomaly detection.

Despite their effectiveness, several supervised OOD methods rely on
graph augmentations or synthetic environments, introducing additional
training complexity and computational overhead. Moreover, representations
learned solely through supervision may become specialized to
label-correlated patterns in the training distribution that do not remain stable under OOD shifts.

\para{Self-Supervised Learning Approaches.}
Self-supervised learning (SSL), particularly graph contrastive learning
(GCL), provides an alternative way to learn graph representations
without label supervision. DGI~\cite{dgi} learns node representations
through local--global mutual information maximization, while
GRACE~\cite{zhu2020grace} contrasts node representations across
augmented graph views. InfoGraph~\cite{sun2019infograph} extends
local--global contrastive learning to graph-level representations.
More recent methods explore GCL in OOD settings. GOOD-D
\cite{goodd_un} targets graph-level OOD detection through local--global contrastive learning with OOD-aware scoring, while DeGEM~\cite{ICLR2025_Decoupled} studies node-level OOD detection on heterophilic graphs using an energy-based model. AD-GCL~\cite{suresh2021adversarial} learns adversarial graph
augmentations to improve representation invariance and incorporates an
information bottleneck to suppress irrelevant information.
MARIO~\cite{mario_unsupervised} studies invariant graph contrastive
learning for node-level OOD generalization under covariate and concept
shifts. Beyond OOD generalization, GCAL~\cite{qiao2025icml-gcal} addresses unsupervised continual adaptation to out-of-distribution graph sequences.

Unlike supervised learning, SSL can capture structural information
without being directly driven by task labels, but its representations
are not explicitly optimized for the downstream prediction task and may
retain task-irrelevant information. This motivates leveraging the
strengths of both paradigms, using SSL representations to complement
label-guided supervised representations. InfoGraph~\cite{sun2019infograph} also combines supervised and
unsupervised objectives for graph representation learning. In contrast,
we study their representational complementarity under OOD and exploit it
through training-time representation fusion and inference-time
prediction fusion.

\subsection{Distance-Based Retrieval Methods for Robust Classification}

Distance-based methods improve OOD robustness by comparing test
representations with training examples and have been widely studied in
vision~\cite{papernot2018deep,park2023nearest,sun2022knnood,adanpc}.
DkNN~\cite{papernot2018deep} estimates prediction confidence from the
consistency of neighboring training representations, while other
non-parametric KNN methods incorporate adaptive neighbor selection or
test-time calibration~\cite{sun2022knnood,adanpc}.
Nearest-neighbor similarity has also been used for post-hoc,
training-free prediction adjustment~\cite{park2023nearest}.
Related ideas have been explored in NLP, where D2U compares model
predictions with a uniform distribution to improve OOD robustness
\cite{yilmaz-toraman-2022-d2u}.

In graph learning, distance-based retrieval remains less explored for
OOD generalization. Our Dual-Space Retrieval framework uses a
non-parametric predictor over supervised and SSL representation spaces,
producing similarity-weighted class scores in each space and combining
them through confidence-aware fusion. This allows complementary
representation spaces to contribute directly at inference time without
training an additional classifier.
\section{Preliminaries}
\label{sec:preliminary}
\subsection{OOD in Graph Structured Data}
% Traditional OOD methods \cite{arora-etal-2021-types, liu2020energy, adanpc, sun2022knnood, lu2021invariant}, primarily focus on Euclidean data, such as images or text, where distribution shifts typically occur in the input feature space. In contrast, graph data presents unique challenges due to its irregular relational structure. Graph-structured data is commonly represented as \((X, A)\), where \(X\) denotes node or edge features and \(A\) is the adjacency matrix. Out-of-distribution scenarios in graphs may involve shifts not only in the feature distribution but also in the graph topology, making the problem inherently more complex.

Distribution shift occurs when the underlying data distribution differs
between training and testing, i.e.,
$P^{\mathrm{train}}(\mathbf{X},\mathbf{Y})
\neq P^{\mathrm{test}}(\mathbf{X},\mathbf{Y})$.
For graph-structured data, such shifts may arise from changes in node
features, graph structure, label distributions, or their relationships.
In this work, we consider graph OOD generalization under naturally
occurring distribution shifts across different domains or time periods.

\subsection{Supervised Graph Representation Learning.}
Given a graph $\mathcal{G}=(\mathbf{X},\mathbf{A},\mathbf{Y})$, where
$\mathbf{X}\in\mathbb{R}^{N\times F}$ is the node feature matrix,
$\mathbf{A}\in\mathbb{R}^{N\times N}$ is the adjacency matrix, and
$\mathbf{Y}$ denotes the node labels, supervised graph representation
learning uses label supervision to learn task-discriminative node
representations. A GNN encoder $f_{\theta}$ maps
$(\mathbf{X},\mathbf{A})$ to node representations $\mathbf{H}$ through
iterative message passing over local neighborhoods.

Let $\mathbf{h}_v^{(k)}$ denote the representation of node $v$ at the
$k$-th GNN layer, with $\mathbf{h}_v^{(0)}=\mathbf{x}_v$. A general
message-passing layer can be expressed as
\begin{equation}
\mathbf{h}_v^{(k)}
=
\operatorname{Upd}^{(k)}
\left(
\mathbf{h}_v^{(k-1)},
\operatorname{Agg}^{(k)}
\left\{
\phi^{(k)}
\left(
\mathbf{h}_v^{(k-1)},
\mathbf{h}_u^{(k-1)},
\mathbf{e}_{uv}
\right)
:u\in\mathcal{N}(v)
\right\}
\right),
\label{eq:gnn}
\end{equation}
where $\mathcal{N}(v)$ denotes the neighbors of node $v$,
$\mathbf{e}_{uv}$ denotes optional edge features,
$\phi^{(k)}$ computes messages from neighboring nodes, and
$\operatorname{Agg}^{(k)}$ and $\operatorname{Upd}^{(k)}$ denote the
aggregation and update functions, respectively.

For node classification, a classification head $c_{\omega}$ maps the
learned representations to node predictions,
$c_{\omega}:\mathbf{H}\rightarrow\widehat{\mathbf{Y}}$.
The encoder and classification head are jointly optimized as
\begin{equation}
(\theta^{*},\omega^{*})
=
\arg\min_{\theta,\omega}
\mathbb{E}_{(\mathbf{X},\mathbf{A},\mathbf{Y})\sim P_{\mathrm{train}}}
\left[
\mathcal{L}_{\mathrm{sup}}
\left(
c_{\omega}\!\left(f_{\theta}(\mathbf{X},\mathbf{A})\right),
\mathbf{Y}
\right)
\right],
\label{eq:sup_objective}
\end{equation}
where $\mathcal{L}_{\mathrm{sup}}$ denotes the cross-entropy loss for
supervised node classification.

\subsection{Self-Supervised Graph Representation Learning.}
\label{sec:preliminary_self}
Unlike supervised representation learning, self-supervised graph
representation learning learns node representations without relying on
node labels. We consider two representative approaches, Deep Graph
Infomax (DGI)~\cite{dgi} and GRACE~\cite{zhu2020grace}, representing distinct
contrastive learning paradigms: \textit{local--global} contrast in DGI
and \textit{node-level cross-view} contrast in GRACE. 

\para{Deep Graph Infomax (DGI).}
DGI learns node representations by maximizing the mutual information
between local node representations and a global summary of the graph.
Given node representations
$\mathbf{H}=f_{\theta}(\mathbf{X},\mathbf{A})$, the global summary
$\mathbf{s}$ is obtained through a readout function
$\mathbf{s}=\mathcal{R}(\mathbf{H})$.
A corruption function $\mathcal{C}$ generates a corrupted graph view
$(\widetilde{\mathbf{X}},\widetilde{\mathbf{A}})
=\mathcal{C}(\mathbf{X},\mathbf{A})$, from which corrupted node
representations
$\widetilde{\mathbf{H}}
=f_{\theta}(\widetilde{\mathbf{X}},\widetilde{\mathbf{A}})$
are obtained. A discriminator $\mathcal{D}$ scores the agreement between
the node representations and the global summary. The DGI objective is
defined as
\begin{equation}
\mathcal{J}_{\mathrm{DGI}}
=
\frac{1}{2N}
\sum_{i=1}^{N}
\left[
\log \mathcal{D}(\mathbf{h}_i,\mathbf{s})
+
\log\left(
1-\mathcal{D}(\widetilde{\mathbf{h}}_i,\mathbf{s})
\right)
\right].
\label{eq:dgi}
\end{equation}
Maximizing this objective encourages high agreement between the original
node representations and the global summary while distinguishing
representations from the corrupted graph view, thereby promoting
local--global mutual information.

\para{GRACE.}
Unlike DGI, which contrasts local node representations with a global graph
summary, GRACE performs node-level contrastive learning across two
independently augmented graph views. Specifically, two stochastic graph
augmentations produce
$(\widetilde{\mathbf{X}}_1,\widetilde{\mathbf{A}}_1)$ and
$(\widetilde{\mathbf{X}}_2,\widetilde{\mathbf{A}}_2)$, which are encoded
by a shared GNN encoder:
\begin{equation}
\mathbf{U}
=
f_{\theta}(\widetilde{\mathbf{X}}_1,\widetilde{\mathbf{A}}_1),
\qquad
\mathbf{V}
=
f_{\theta}(\widetilde{\mathbf{X}}_2,\widetilde{\mathbf{A}}_2).
\label{eq:grace_views}
\end{equation}
For each node, its representations across the two views form a positive
pair, while representations of other nodes serve as negatives. GRACE
maximizes the following symmetric node-level InfoNCE objective:
\begin{equation}
\mathcal{J}_{\mathrm{GRACE}}
=
\frac{1}{2N}
\sum_{i=1}^{N}
\left[
\ell(\mathbf{u}_i,\mathbf{v}_i)
+
\ell(\mathbf{v}_i,\mathbf{u}_i)
\right],
\label{eq:grace}
\end{equation}
where $\ell(\cdot,\cdot)$ denotes the InfoNCE score that encourages
agreement between positive pairs relative to negative pairs.

\noindent\textbf{Downstream Node Classification.}
After self-supervised training with either DGI or GRACE, the learned GNN
encoder is frozen and used to extract node representations for downstream
classification. A trainable linear classifier $c_{\omega}$ is then learned
on the labeled training nodes:
\begin{equation}
\omega^{*}
=
\arg\min_{\omega}
\mathbb{E}_{(\mathbf{X},\mathbf{A},\mathbf{Y})\sim P_{\mathrm{train}}}
\left[
\mathcal{L}_{\mathrm{sup}}
\left(
c_{\omega}
\left(
f_{\theta^{*}}(\mathbf{X},\mathbf{A})
\right),
\mathbf{Y}
\right)
\right],
\label{eq:ssl_downstream}
\end{equation}
where $\theta^{*}$ denotes the frozen encoder parameters learned by the
self-supervised objective.

% defnition: ood in graph. temepral time shift. 
% spatial shift: social network...node, graph... all the datasets, feature, graph struture. 

% Compared with baseline method~\cite{wu2022eerm} and ~\cite{andyZhouKG}

% Main idea: use unsupervised to augment supervised. 

% 1. Align supervised and unsupervised
% 2. Majority vote

% Classifier-based:

% Classifier-free: weighted majority vote. GNN embeddings closet match based on embedding distance because it contains both feature and structure information for proximity. 

% majority vote: self consistency
\section{Methodology}
\label{sec:method}
\begin{figure}[t]
  \centering
  \includegraphics[width=\columnwidth]{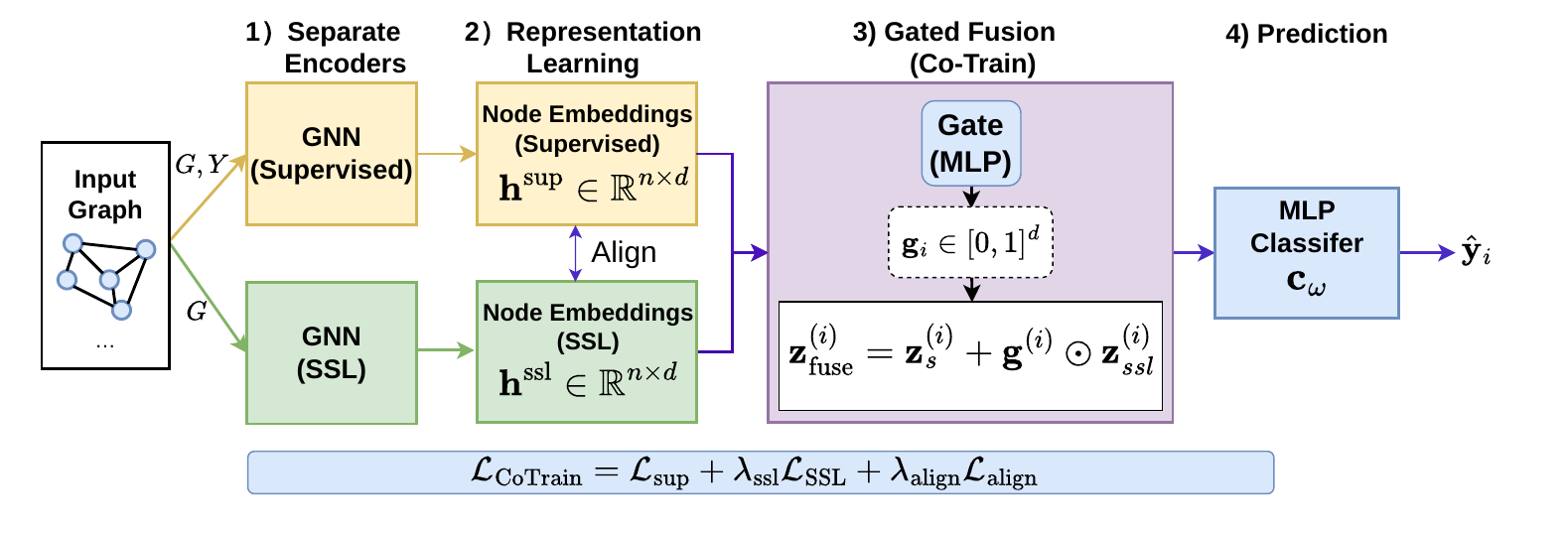}
  \caption{Overview of Co-Train. The supervised and SSL branches use
separate encoders without parameter sharing and are jointly optimized.
A learned gate adaptively incorporates the complementary SSL
representation into the supervised representation for prediction.}
  \label{fig:Co-train-framework}
\end{figure}

\begin{figure}[t]
  \centering
  \includegraphics[width=\columnwidth]{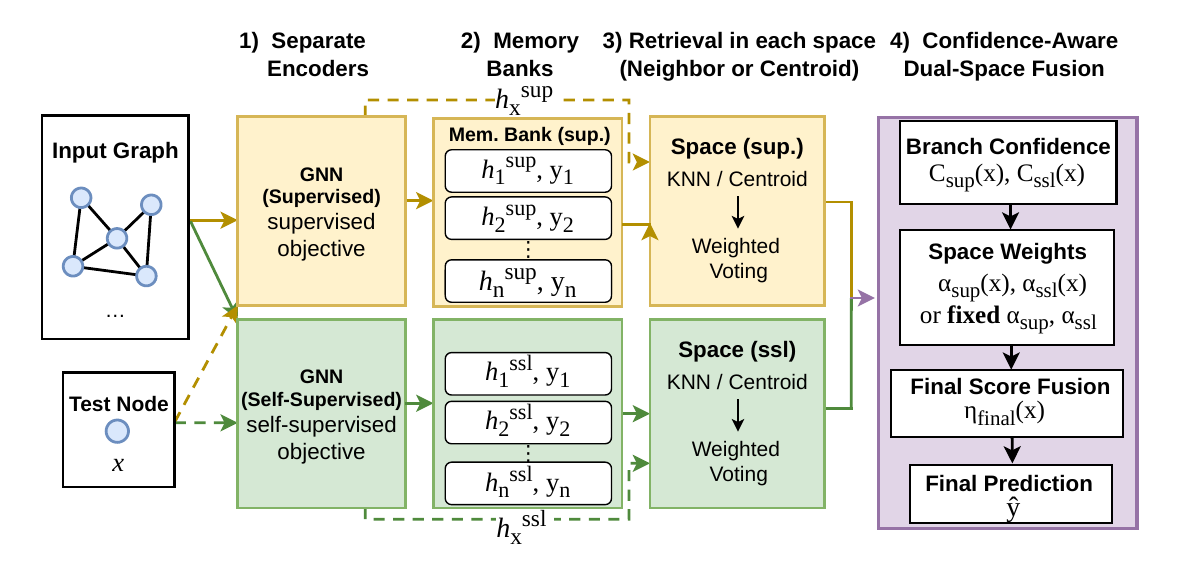}
  \caption{Overview of the Dual-Space Retrieval framework.
Supervised and self-supervised GNNs construct separate embedding
memory banks. At test time, neighbor- or centroid-based retrieval
produces similarity-weighted prediction scores in each space, which
are combined through confidence-aware dual-space score fusion.}
  \label{fig:Retreival_framework}
\end{figure}

Our framework uses supervised representations as the primary prediction
space and SSL representations as complementary signals for OOD
generalization. We develop two mechanisms, illustrated in
Figures~\ref{fig:Co-train-framework} and~\ref{fig:Retreival_framework}.
\textbf{Co-Train} jointly optimizes and integrates two separately
parameterized encoders during training, whereas \textbf{Dual-Space
Retrieval} uses a non-parametric predictor over independently learned
representations and combines their prediction scores at inference time
through confidence-aware fusion. Both frameworks are backbone-agnostic
and allow the supervised and SSL branches to use different GNN
architectures.

\subsection{Co-Train}
\label{sec:cotrain}
Our Co-Train framework jointly learns task-supervised and self-supervised
representations using two GNN encoders. Given an input graph
$(\mathbf{X}, \mathbf{A})$, the supervised encoder $f_{\theta_s}$ learns
representations guided by node labels, while the self-supervised encoder
$f_{\theta_u}$ captures complementary structural information through a
self-supervised objective. Importantly, the two encoders \textit{do not
share parameters} and are \textit{not required to use the same backbone
architecture}, allowing the two branches to learn representations under
their respective objectives.

We instantiate the self-supervised branch with either DGI or GRACE,
representing local--global and node-level cross-view contrastive learning,
respectively. This allows us to examine whether the benefits of Co-Train
generalize across distinct self-supervised objectives.

For node $i$, let $\mathbf{h}^{(i)}_s$ and $\mathbf{h}^{(i)}_ssl$ denote
the representations produced by the supervised and self-supervised
encoders, respectively. We first map them into a common $d$-dimensional
space using two \textit{learnable projection} functions:
\begin{equation}
\mathbf{z}^{(i)}_s
=
p_s(\mathbf{h}^{(i)}_s),
\qquad
\mathbf{z}^{(i)}_{ssl}
=
p_{ssl}(\mathbf{h}^{(i)}_{ssl}).
\label{eq:cotrain_projection}
\end{equation}

To combine the supervised and self-supervised representations, we use
a gated fusion mechanism. Rather than directly adding the two
representations, we compute an embedding-dependent gate from both
representations to adaptively weight the SSL embedding before fusion.
Let $\bar{\mathbf{z}}^{(i)}_s$ and
$\bar{\mathbf{z}}^{(i)}_{ssl}$ denote the $\ell_2$-normalized projected
representations. The gate is computed as
\begin{equation}
\mathbf{g}^{(i)} =
\sigma \left(
f_g \left(
\left[
\bar{\mathbf{z}}^{(i)}_s
\Vert
\bar{\mathbf{z}}^{(i)}_{ssl}
\right]
\right)
\right),
\label{eq:cotrain_gate}
\end{equation}
where $\Vert$ denotes concatenation, $f_g$ is a learnable gating
network, and $\sigma(\cdot)$ is the sigmoid function. The final
representation is
\begin{equation}
\mathbf{z}^{(i)}_{\mathrm{fuse}}
=
\mathbf{z}^{(i)}_s
+
\mathbf{g}^{(i)} \odot \mathbf{z}^{(i)}_{ssl},
\label{eq:cotrain_fusion}
\end{equation}
where $\odot$ denotes element-wise multiplication. This keeps the
supervised representation as the base while adaptively controlling
the contribution of the SSL representation. We compare gated fusion
with direct addition and concatenation in Section~\ref{sec:ablation}.

The fused representation $\mathbf{z}^{(i)}_{\mathrm{fuse}}$ is fed into
the supervised classification head $c_\omega$, and the resulting
predictions are optimized using the supervised loss
$\mathcal{L}_{\mathrm{sup}}$ defined in Eq.~\ref{eq:sup_objective}. In addition, we encourage the supervised and self-supervised representations to remain compatible
through an alignment objective over their normalized projected embeddings:
\begin{equation}
    \mathcal{L}_{\mathrm{align}}
    =
    \operatorname{MSE}
    \left(
        \bar{\mathbf{Z}}_s,
        \bar{\mathbf{Z}}_{ssl}
    \right).
\end{equation}

The self-supervised branch is jointly optimized using
$\mathcal{L}_{\mathrm{SSL}}$, which denotes the minimization form of the
corresponding DGI or GRACE objective defined in Section~\ref{sec:preliminary_self}; that is,
$\mathcal{L}_{\mathrm{SSL}}=-\mathcal{J}_{\mathrm{DGI}}$ for DGI and
$\mathcal{L}_{\mathrm{SSL}}=-\mathcal{J}_{\mathrm{GRACE}}$ for GRACE.
The overall Co-Train objective is then
\begin{equation}
    \mathcal{L}_{\mathrm{CoTrain}}
    =
    \mathcal{L}_{\mathrm{sup}}
    +
    \lambda_{\mathrm{ssl}}\mathcal{L}_{\mathrm{SSL}}
    +
    \lambda_{\mathrm{align}}\mathcal{L}_{\mathrm{align}},
\end{equation}
where $\lambda_{\mathrm{ssl}}$ and $\lambda_{\mathrm{align}}$ control
the contributions of the self-supervised and alignment objectives,
respectively.

% \paragraph{Differences from InfoGraph}
% While our framework shares a similar joint supervised-unsupervised training structure with InfoGraph \cite{sun2019infograph}, it differs significantly in both objective and design. InfoGraph is designed for graph-level classification and introduces an enhanced variant of DGI for that purpose. In contrast, we focus on node classification and demonstrates that even with the original DGI, when combined with supervised learning via feature fusion and joint training, can significantly improve \textit{out-of-distribution detection} at the node level. Moreover, whereas InfoGraph maximizes the similarity between supervised and unsupervised embeddings through an additional alignment loss, we directly fuse the two representations. This simple integration proves to be effective without requiring explicit alignment constraints. \qy{not sure if it is clear enough}
% \subsection{Retrieval-Vote Ensemble} 
% \qy{intuition: maybe adanpc?}
% Building on our classifier-based approach that uses joint encoder training and feature fusion, we now explore a non-parametric, distance-based method. Inspired by recent success of classifier-free methods in out-of-detection and domain adaptation, we propose a \textit{retrieval-based} approach that performs \textit{weighted majority voting} over the \({k}\)-nearest neighbors, using embeddings from both supervised and unsupervised GNN models. 
\subsection{Dual-Space Retrieval}
\para{Dual-Space Memory Construction.}
Given a set of labeled training nodes
$\mathcal{D}_{\mathrm{tr}}=\{(x_i,y_i)\}_{i=1}^{N}$,
we obtain the supervised and self-supervised representations following
the learning objectives defined in
Equations~\ref{eq:sup_objective} and~\ref{eq:ssl_downstream},
respectively.
Let $\mathbf{h}_{\mathrm{sup}}^{(i)}$ and
$\mathbf{h}_{\mathrm{ssl}}^{(i)}$ denote the representations of node $i$
obtained from the supervised and self-supervised encoders, respectively.
We construct two memory banks, one for each representation space:
\begin{equation}
    \mathcal{M}_{\mathrm{sup}}
    =
    \left\{
        \left(\mathbf{h}_{\mathrm{sup}}^{(i)}, y_i\right)
    \right\}_{i=1}^{N},
    \qquad
    \mathcal{M}_{\mathrm{ssl}}
    =
    \left\{
        \left(\mathbf{h}_{\mathrm{ssl}}^{(i)}, y_i\right)
    \right\}_{i=1}^{N}.
    \label{eq:memory_bank}
\end{equation}
The two memory banks preserve the supervised and self-supervised
representation spaces independently, with each training representation
paired with its corresponding class label. Importantly, the two encoders are independently parameterized and need not share the same backbone architecture.

\para{Neighbor-Based Retrieval}
Given a test node $x$, we perform retrieval independently in the
supervised and self-supervised representation spaces. For each branch
$b\in\{\mathrm{sup},\mathrm{ssl}\}$, we retrieve the $k$ nearest
training nodes from its corresponding memory bank $\mathcal{M}_b$
according to cosine similarity. Specifically, the similarity between
$x$ and a retrieved node $j$ is computed as
\begin{equation}
s_{xj}^{b}
=
\frac{
    {\mathbf{h}_{b}^{(x)}}^\top \mathbf{h}_{b}^{(j)}
}{
    \|\mathbf{h}_{b}^{(x)}\|_2
    \|\mathbf{h}_{b}^{(j)}\|_2
}.
\label{eq:retrieval_similarity}
\end{equation}

Within each representation space, we use the similarity-weighted
non-parametric voting scheme of AdaNPC~\cite{adanpc} to obtain
prediction scores from the retrieved neighbors. Their cosine
similarities are normalized into voting weights:
\begin{equation}
w_{xj}^{b}
=
\frac{
    \exp(s_{xj}^{b}/\tau_r)
}{
    \sum_{\ell\in\mathcal{N}_k(x,\mathcal{M}_b)}
    \exp(s_{x\ell}^{b}/\tau_r)
},
\label{eq:retrieval_weight}
\end{equation}
where $\tau_r$ is the retrieval temperature controlling the sharpness
of the neighbor weights. We then aggregate the one-hot labels of the
retrieved neighbors using their similarity weights:
\begin{equation}
\boldsymbol{\eta}_{b}(x)
=
\sum_{j\in\mathcal{N}_k(x,\mathcal{M}_b)}
w_{xj}^{b}\mathbf{y}_{j},
\qquad
b\in\{\mathrm{sup},\mathrm{ssl}\},
\label{eq:retrieval_score}
\end{equation}
where $\mathbf{y}_{j}\in\{0,1\}^{C}$ denotes the one-hot class label
of neighbor $j$. Applying this procedure independently to the two
memory banks yields the class-wise prediction score vectors
$\boldsymbol{\eta}_{\mathrm{sup}}(x)$ and
$\boldsymbol{\eta}_{\mathrm{ssl}}(x)$ from the supervised and
self-supervised representation spaces, respectively.

% \para{Class-Centroid Retrieval}
% For multiclass datasets, we additionally consider a class-centroid
% variant that provides an explicit representative for each class.
% Unlike instance-level neighbor retrieval over individual training
% nodes, class-level centroid retrieval summarizes each class with a
% single representation and performs retrieval over these class
% representatives.

% For each branch $b\in\{\mathrm{sup},\mathrm{ssl}\}$ and class $c$, we
% compute the class centroid as
% \begin{equation}
% \boldsymbol{\mu}_{b,c}
% =
% \frac{1}{|\mathcal{D}_c|}
% \sum_{i\in\mathcal{D}_c}
% \mathbf{h}_{b}^{(i)},
% \label{eq:class_centroid}
% \end{equation}
% where $\mathcal{D}_c$ denotes the set of training nodes belonging to
% class $c$. The $C$ resulting centroids form a class-level memory bank
% for each representation branch.

% For a test node $x$, we retrieve the top-$k$ class centroids and
% obtain class-wise prediction scores using the same similarity-weighted
% voting procedure in
% Equations~\ref{eq:retrieval_similarity}--\ref{eq:retrieval_score},
% with each centroid carrying its corresponding class label.

% This variant is particularly suitable when embeddings within each
% class are relatively compact, allowing a class centroid to provide a
% meaningful class-level representation. It also provides a compact
% alternative to instance-level neighbor retrieval for multiclass
% prediction.
% \qy{We further evaluate a graph-aware feature-smoothing variant, de-
% noted as Retrieval*, which averages each node embedding with its
% graph neighbors before retrieval.}

\para{Class-Centroid Retrieval.}
For multiclass prediction, instance-level KNN can be dominated by a
small set of locally retrieved classes. We therefore consider
class-centroid retrieval, which represents each class explicitly in
the retrieval space.

For each branch $b\in\{\mathrm{sup},\mathrm{ssl}\}$ and class $c$, we
compute the class centroid as
\begin{equation}
\boldsymbol{\mu}_{b,c}
=
\frac{1}{|\mathcal{D}_c|}
\sum_{i\in\mathcal{D}_c}
\mathbf{h}_{b}^{(i)},
\label{eq:class_centroid}
\end{equation}
where $\mathcal{D}_c$ denotes the set of training nodes belonging to
class $c$. The $C$ centroids form a class-level memory bank for each
representation space. For a test node $x$, we perform retrieval over
the class centroids using the same similarity-weighted scoring in
Equations~\ref{eq:retrieval_similarity}--\ref{eq:retrieval_score}.

Additionally, we apply graph-aware feature smoothing before centroid
construction, where each node representation is averaged with those of
its graph neighbors. This reduces local representation variation and
encourages neighboring nodes to form more coherent representations,
yielding more representative class centroids for retrieval.

\para{Confidence-Aware Dual-Space Fusion.}
The two representation spaces may provide prediction signals of
different strengths for the same test node. We therefore adaptively
determine their relative contributions based on retrieval confidence,
assigning a larger weight to the branch with a more confident prediction.
For each branch, we define its confidence as
\begin{equation}
C_b(x)
=
\max_c [\boldsymbol{\eta}_b(x)]_c,
\qquad
b\in\{\mathrm{sup},\mathrm{ssl}\}.
\label{eq:retrieval_confidence}
\end{equation}
The contribution of the supervised branch is then
\begin{equation}
\alpha_{\mathrm{sup}}(x)
=
\frac{
    \exp(C_{\mathrm{sup}}(x)/\tau_f)
}{
    \exp(C_{\mathrm{sup}}(x)/\tau_f)
    +
    \exp(C_{\mathrm{ssl}}(x)/\tau_f)
},
\label{eq:fusion_weight}
\end{equation}
with
$\alpha_{\mathrm{ssl}}(x)=1-\alpha_{\mathrm{sup}}(x)$.

We further account for the agreement between the two prediction
signals. Specifically, we measure their hard-label disagreement on
the unlabeled target nodes $\mathcal{X}_t$ as
% \begin{equation}
% D =
% \frac{1}{|\mathcal{X}_t|}
% \sum_{x\in\mathcal{X}_t}
% \mathbbm{1}
% [\hat y_{\mathrm{sup}}(x)\neq\hat y_{\mathrm{ssl}}(x)],
% \quad
% \hat y_b(x)=\arg\max_c[\boldsymbol{\eta}_b(x)]_c .
% \label{eq:branch_disagreement}
% \end{equation}
\begin{equation}
\delta =
\frac{1}{|\mathcal{X}_t|}
\sum_{x\in\mathcal{X}_t}
\mathbb{I}[\hat{y}_{\mathrm{sup}}(x) \neq
\hat{y}_{\mathrm{ssl}}(x)],
\quad
\hat y_b(x)=\arg\max_c[\boldsymbol{\eta}_b(x)]_c .
\label{eq:branch_disagreement}
\end{equation}

When the two branches exhibit large disagreement ($\delta>0.5$), we
place greater emphasis on the supervised prediction by setting
$\alpha_{\mathrm{sup}}=1$ and using a small fixed
$\alpha_{\mathrm{ssl}}$ (e.g., $0.01$), instead
of the confidence-based weights. Otherwise, we use the confidence-based
weights defined above.

Finally, we combine the two prediction scores as
\begin{equation}
\boldsymbol{\eta}_{\mathrm{final}}(x)
=
\alpha_{\mathrm{sup}}(x)\boldsymbol{\eta}_{\mathrm{sup}}(x)
+
\alpha_{\mathrm{ssl}}(x)\boldsymbol{\eta}_{\mathrm{ssl}}(x),
\label{eq:retrieval_fusion}
\end{equation}
and predict
$\hat y=\arg\max_c[\boldsymbol{\eta}_{\mathrm{final}}(x)]_c$.

\section{Experiments}
\label{sec:experiments}

\definecolor{lightblue}{rgb}{0.8, 0.88, 1}
\definecolor{midblue}{rgb}{0.4, 0.6, 1}
\definecolor{darkblue}{rgb}{0.1, 0.1, 0.8}

\definecolor{lightred}{rgb}{1, 0.8, 0.8}
\definecolor{midred}{rgb}{1, 0.4, 0.4}
\definecolor{darkred}{rgb}{0.8, 0.0, 0.0}
\definecolor{tan}{rgb}{0.82, 0.71, 0.55}
\subsection{Experiment Setups}

\begin{table*}[t]
\centering
\normalsize
\setlength{\tabcolsep}{1.5pt}
\caption{F1-score (\%) on \texttt{Elliptic} across nine temporal test
splits and different GNN backbones.}
\label{tab:elliptic-results}
\begin{tabular}{l|ccccccccc}
\toprule
\textbf{Method (Backbone)} & T1 & T2 & T3 & T4 & T5 & T6 & T7 & T8 & T9\\
\midrule
ERM (SAGE)
& 91.74 $\pm$ 1.85 & 85.27 $\pm$ 1.39 & 77.42 $\pm$ 1.41 &
73.64 $\pm$ 1.45 & 75.24 $\pm$ 2.88 & 76.92 $\pm$ 4.81 &
80.68 $\pm$ 1.74 & 65.96 $\pm$ 3.97 & 49.03 $\pm$ 0.85 \\

EERM (SAGE)
& 86.59 $\pm$ 2.71 & 81.03 $\pm$ 2.55 & 76.68 $\pm$ 1.39 &
72.24 $\pm$ 1.99 & 75.20 $\pm$ 2.20 & 78.28 $\pm$ 3.33 &
72.77 $\pm$ 4.73 & 63.83 $\pm$ 3.98 & 49.14 $\pm$ 1.01 \\

LiSA (SAGE)
& 92.58 $\pm$ 7.91 & 85.23 $\pm$ 8.41 & \cellcolor{brown!20}{79.00 $\pm$ 4.70} &
74.45 $\pm$ 5.10 & 74.95 $\pm$ 1.75 & 68.46 $\pm$ 4.90 &
70.42 $\pm$ 2.74 & 55.12 $\pm$ 2.09 & 47.19 $\pm$ 0.92 \\

MARIO (SAGE)
& 90.51 $\pm$ 4.12 & 84.73 $\pm$ 2.02 & 77.62 $\pm$ 2.45 &
72.05 $\pm$ 3.42 & 73.18 $\pm$ 4.65 & 77.35 $\pm$ 6.43 &
63.71 $\pm$ 8.36 & 65.02 $\pm$ 7.16 & 48.79 $\pm$ 0.03 \\
DGI (SAGE) & {56.39 $\pm$ 6.63} & {53.46 $\pm$ 5.17} & {55.13 $\pm$ 5.19} & {52.81 $\pm$ 6.09} & {59.09 $\pm$ 5.01} & {56.12 $\pm$ 4.01} & {57.98 $\pm$ 4.77} & {50.36 $\pm$ 1.54} & {49.88 $\pm$ 1.25} \\
GRACE (SAGE) & {90.07 $\pm$ 3.14} & {83.72 $\pm$ 1.35} & {78.11 $\pm$ 1.72} & {72.22 $\pm$ 2.13} & {75.15 $\pm$ 3.25} & {80.58 $\pm$ 2.98} & {77.34 $\pm$ 5.95} & \cellcolor{tan}{71.03 $\pm$ 4.09} & {48.70 $\pm$ 0.08} \\
CoTrain-DGI (SAGE)
& \cellcolor{brown!20}{93.78 $\pm$ 1.61}
& \cellcolor{tan}{86.60 $\pm$ 1.07}
& \cellcolor{tan}{79.49 $\pm$ 1.32}
& \cellcolor{tan}{74.98 $\pm$ 0.76}
& \cellcolor{tan}{77.86 $\pm$ 1.57}
& \cellcolor{tan}{85.07 $\pm$ 1.27}
& \cellcolor{brown!20}{85.22 $\pm$ 0.75}
& \cellcolor{brown!20}{70.37 $\pm$ 1.56}
& {49.91 $\pm$ 0.24} \\
CoTrain-GRACE (SAGE)
& \cellcolor{tan}{93.98 $\pm$ 1.39}
& \cellcolor{brown!20}{85.40 $\pm$ 0.96}
& {77.57 $\pm$ 1.34}
& {71.98 $\pm$ 1.58}
& \cellcolor{brown!20}{75.91 $\pm$ 1.85}
& \cellcolor{brown!20}{84.50 $\pm$ 1.67}
& \cellcolor{tan}{85.72 $\pm$ 1.14}
& {70.33 $\pm$ 1.34}
& \cellcolor{brown!20}{49.95 $\pm$ 0.48} \\
% Retrieval-DGI (SAGE) & {91.55 $\pm$ 1.12} & {84.08 $\pm$ 0.85} & {78.17 $\pm$ 0.72} & \cellcolor{brown!20}{74.83 $\pm$ 1.02} & {72.29 $\pm$ 2.30} & {80.04 $\pm$ 1.35} & {77.49 $\pm$ 1.60} & {66.06 $\pm$ 1.80} & \cellcolor{tan}{51.44 $\pm$ 0.75} \\
% Retrieval-GRACE (SAGE)
% & {93.16 $\pm$ 1.02}
% & \cellcolor{brown!20}{85.40 $\pm$ 0.96}
% & {78.79 $\pm$ 0.71}
% & {74.70 $\pm$ 0.79}
% & {73.09 $\pm$ 0.75}
% & {81.12 $\pm$ 1.04}
% & {77.10 $\pm$ 1.67}
% & {66.83 $\pm$ 1.53}
% & {49.80 $\pm$ 0.84} \\
Retrieval-DGI (SAGE)
& {89.83 $\pm$ 1.25}
& {82.82 $\pm$ 2.04}
& {77.31 $\pm$ 2.33}
& {71.69 $\pm$ 6.01}
& {69.40 $\pm$ 4.60}
& {77.68 $\pm$ 2.69}
& {71.73 $\pm$ 3.23}
& {66.09 $\pm$ 3.11}
& {50.22 $\pm$ 0.80} \\

Retrieval-GRACE (SAGE)
& {92.03 $\pm$ 1.42}
& {84.62 $\pm$ 0.74}
& {78.35 $\pm$ 0.71}
& \cellcolor{brown!20}{74.45 $\pm$ 0.65}
& {72.94 $\pm$ 0.93}
& {80.64 $\pm$ 0.88}
& {77.13 $\pm$ 0.87}
& {66.70 $\pm$ 1.20}
& \cellcolor{tan}{50.26 $\pm$ 0.85} \\
\midrule
ERM (GCNII) & 78.89 $\pm$ 3.51 & 71.00 $\pm$ 4.61 & 64.37 $\pm$ 5.72 &
63.65 $\pm$ 7.81 & 66.48 $\pm$ 5.92 & 67.72 $\pm$ 5.72 &
72.11 $\pm$ 5.03 & 59.43 $\pm$ 3.36 & 49.04 $\pm$ 1.01 \\

EERM (GCNII) & 61.27 $\pm$ 2.81 & 66.99 $\pm$ 2.49 & 63.66 $\pm$ 2.20 &
61.59 $\pm$ 2.60 & 60.49 $\pm$ 4.61 & 57.37 $\pm$ 4.96 &
56.58 $\pm$ 3.92 & 48.17 $\pm$ 5.77 &\cellcolor{tan}{50.98 $\pm$ 4.88} \\

LiSA (GCNII) & 77.26 $\pm$ 5.87 & 73.81 $\pm$ 4.35 & 68.75 $\pm$ 4.56 &
67.18 $\pm$ 3.64 & 66.99 $\pm$ 4.18 & 67.95 $\pm$ 7.00 &
66.46 $\pm$ 5.59 & 55.99 $\pm$ 6.82 &\cellcolor{brown!20}{50.27 $\pm$ 3.17} \\

MARIO (GCNII) & 93.72 $\pm$ 2.12 & 86.92 $\pm$ 0.91 & 79.80 $\pm$ 1.25 &
\cellcolor{tan}{75.82 $\pm$ 2.49}& 77.88 $\pm$ 3.50 & 80.52 $\pm$ 4.85 &
71.41 $\pm$ 10.13 & 65.17 $\pm$ 5.30 & 48.88 $\pm$ 0.46 \\
DGI (GCNII) & {55.90 $\pm$ 2.79} & {55.50 $\pm$ 5.44} & {50.82 $\pm$ 3.70} & {50.96 $\pm$ 7.11} & {49.18 $\pm$ 3.69} & {52.20 $\pm$ 3.11} & {50.98 $\pm$ 3.97} & {49.13 $\pm$ 1.26} & {49.01 $\pm$ 2.00} \\
GRACE (GCNII) & {72.42 $\pm$ 4.88} & {72.83 $\pm$ 4.22} & {66.81 $\pm$ 5.82} & {65.64 $\pm$ 5.01} & {66.41 $\pm$ 6.87} & {64.78 $\pm$ 6.11} & {62.62 $\pm$ 4.63} & {57.75 $\pm$ 5.35} & {48.76 $\pm$ 0.41} \\
CoTrain-DGI (GCNII)
& \cellcolor{brown!20}{95.31 $\pm$ 0.62}
& \cellcolor{tan}{87.96 $\pm$ 0.80}
& \cellcolor{brown!20}{79.81 $\pm$ 0.93}
& {74.74 $\pm$ 0.56}
& \cellcolor{brown!20}{77.90 $\pm$ 2.07}
& \cellcolor{brown!20}{86.37 $\pm$ 1.08}
& \cellcolor{brown!20}{87.47 $\pm$ 1.34}
& \cellcolor{brown!20}{72.72 $\pm$ 2.97}
& {49.77 $\pm$ 0.40} \\
CoTrain-GRACE (GCNII)
& \cellcolor{tan}{95.39 $\pm$ 0.43}
& \cellcolor{brown!20}{87.87 $\pm$ 0.52}
& \cellcolor{tan}{80.17 $\pm$ 0.77}
& \cellcolor{brown!20}{74.93 $\pm$ 0.58}
& \cellcolor{tan}{79.07 $\pm$ 1.34}
& \cellcolor{tan}{87.61 $\pm$ 0.75}
& \cellcolor{tan}{87.63 $\pm$ 0.79}
& \cellcolor{tan}{75.61 $\pm$ 1.05}
& {49.83 $\pm$ 0.31} \\
% Retrieval-DGI (GCNII) & {80.16 $\pm$ 6.87} & {69.66 $\pm$ 5.62} & {65.33 $\pm$ 5.86} & {66.63 $\pm$ 6.14} & {71.41 $\pm$ 2.82} & {76.29 $\pm$ 2.85} & {77.22 $\pm$ 1.75} & {62.81 $\pm$ 2.06} & \cellcolor{tan}{49.33 $\pm$ 0.85} \\
% Retrieval-GRACE (GCNII) & {87.09 $\pm$ 7.66} & {79.08 $\pm$ 7.63} & {72.30 $\pm$ 7.57} & {67.96 $\pm$ 8.12} & {73.72 $\pm$ 7.96} & {80.69 $\pm$ 7.40} & {80.23 $\pm$ 8.26} & \cellcolor{tan}{75.85 $\pm$ 5.46} & {48.75 $\pm$ 0.52} \\
Retrieval-DGI (GCNII)
& {82.29 $\pm$ 3.31}
& {70.40 $\pm$ 3.50}
& {65.54 $\pm$ 3.93}
& {67.39 $\pm$ 5.56}
& {70.85 $\pm$ 2.87}
& {75.25 $\pm$ 3.37}
& {77.84 $\pm$ 2.18}
& {62.84 $\pm$ 2.00}
& {49.32 $\pm$ 0.92} \\

Retrieval-GRACE (GCNII)
& {83.91 $\pm$ 4.96}
& {74.57 $\pm$ 5.12}
& {70.19 $\pm$ 5.28}
& {70.62 $\pm$ 4.45}
& {73.11 $\pm$ 2.12}
& {75.57 $\pm$ 2.82}
& {79.21 $\pm$ 1.71}
& {63.67 $\pm$ 2.10}
& {49.30 $\pm$ 0.88} \\

\bottomrule
\end{tabular}
\end{table*}

\para{Datasets.}
We evaluate OOD node classification under two natural distribution
shifts: temporal shifts in dynamic graphs and domain shifts across
multiple graphs. Full dataset details are provided in Table~\ref{tab:datasets} in Sec.~\ref{sec:app-data}.

\begin{itemize}[leftmargin=*, noitemsep, topsep=0pt]
\item \textbf{Temporal transfer.}
We use \texttt{Elliptic}~\cite{elliptic_dataset} for illicit transaction
classification across time, and \texttt{OGB-Arxiv}~\cite{ogb-arxiv} for
paper subject classification across publication years.

\item \textbf{Cross-domain transfer.}
We use \texttt{Twitch-Explicit (Twitch)}~\cite{twitch_dataset} for explicit-language
classification across regional user interaction graphs, and
\texttt{Facebook-100 (FB-100)}~\cite{fb100_dataset} for gender classification
across university social networks.
\end{itemize}
\para{Baselines.}
We compare our method against representative supervised and
self-supervised baselines:
\begin{itemize}[leftmargin=*, noitemsep, topsep=0pt]
    \item \textbf{ERM}~\cite{erm}: A standard supervised learning
    baseline based on empirical risk minimization.
    
    \item \textbf{EERM}~\cite{wu2022eerm}: A supervised OOD method that
    constructs augmented environments via reinforcement learning for
    invariant learning.
    
    \item \textbf{DGI}~\cite{dgi}: A self-supervised graph representation
    learning method based on local--global mutual information maximization.
    
    \item \textbf{GRACE}~\cite{zhu2020grace}: A self-supervised graph
    contrastive learning method based on augmented graph views.
    
    \item \textbf{LiSA}~\cite{lisa}: A supervised OOD method that learns
    invariant representations using label-invariant subgraphs and
    energy-based regularization.
    
    \item \textbf{MARIO}~\cite{mario_unsupervised}: A self-supervised
    graph OOD method combining adversarial graph augmentation and
    information bottleneck regularization.
\end{itemize}
For fair comparison, we follow the same OOD data splits used in
EERM and LiSA for final evaluation.

\para{Implementation Details.}
For each dataset, we evaluate our methods with different supervised
GNN backbones to examine generalizability across model architectures.
Under each setting, all baselines use the same backbone as the
supervised branch of our methods for a fair comparison. For the SSL
branch, we use either GCN or GraphSAGE based on validation performance.
DGI and GRACE use the same SSL encoder architecture within each setting
but are trained separately. Once the supervised and SSL architectures are selected, we use the same backbone configuration for Co-Train and Dual-Space Retrieval, enabling a controlled evaluation across different backbone architectures. Results with alternative
SSL encoder architectures are provided in Sec.~\ref{sec:app-ssl-encoder-sensitivity}. 

For Dual-Space Retrieval, we use $k=30$, $100$, $500$, and $300$ for
\texttt{OGB-Arxiv}, \texttt{Elliptic}, \texttt{Twitch}, and
\texttt{FB100}, respectively, selected based on validation performance.
The same $k$ is used for the supervised and SSL retrieval spaces.
Sensitivity to $k$ is reported in
Sec.~\ref{sec:app-sensitive_k}. We fix the retrieval
similarity temperature $\lambda_r$ to $0.01$ and the confidence-fusion
temperature $\tau_f$ to $0.5$ across all datasets. When the two branches
exhibit large disagreement ($\delta>0.5$), we reduce the SSL contribution
using a small fixed weight $\alpha_{\mathrm{ssl}}$. This occurs on
\texttt{OGB-Arxiv}, where we set $\alpha_{\mathrm{ssl}}=0.01$.
Sensitivity to $\alpha_{\mathrm{ssl}}$ is reported in
Table~\ref{tab:arxiv-epsilon-ablation} in Sec.~\ref{sec:app-sensitive_fusion_weight}.

For Co-Train, we set $\lambda_{\mathrm{align}}=0.02$ and warm up
$\lambda_{\mathrm{ssl}}$ from $0.1$ to $0.7$ during training. All
reported performance is evaluated on the test sets and averaged over
10 runs, with mean and standard deviation reported.

\subsection{Temporal Generalization Performance}
We evaluate temporal generalization under two settings: a dynamic graph
(\texttt{Elliptic}) that evolves across time snapshots and a static graph
(\texttt{OGB-Arxiv}) whose nodes are partitioned by publication time.
These settings capture distinct forms of temporal distribution shift.
In both cases, incorporating SSL representations benefits supervised OOD
generalization, with particularly consistent gains from Co-Train.

\para{Temporal Generalization on Dynamic Graphs.}
We evaluate temporal distribution shifts on the Elliptic transaction
network. We use graph snapshots 7--11 for training and snapshots
17--49 for testing, chronologically grouped into nine evaluation
periods. Table~\ref{tab:elliptic-results} reports F1 scores for illicit
transaction classification. We consider GraphSAGE and GCNII as the
supervised backbones, paired with GCN and GraphSAGE encoders,
respectively, for the SSL branch.

Co-Train achieves strong and consistent performance across both
backbone settings. With GraphSAGE, CoTrain-DGI and CoTrain-GRACE
achieve average F1 scores of 78.14\% and 77.26\%, outperforming the
strongest baseline on average (GRACE, 75.21\%) by 2.93 and 2.05
percentage points, respectively. With GCNII, the gains are even more
pronounced: CoTrain-DGI and CoTrain-GRACE achieve average F1 scores
of 79.12\% and 79.79\%, exceeding the strongest baseline (MARIO,
75.57\%) by 3.55 and 4.22 points. The improvements are sustained
across the temporal sequence; for example, at T7, CoTrain-GRACE
reaches 85.72\% and 87.63\% with GraphSAGE and GCNII, compared with
the strongest baselines of 80.68\% and 71.41\%, respectively.

Dual-Space Retrieval is also competitive, particularly in later
periods. With GraphSAGE, Retrieval-GRACE achieves an average F1 of
75.24\%, comparable to the strongest baseline (75.21\%). With GCNII,
although retrieval is less competitive on average, Retrieval-GRACE
reaches 79.21\% at T7, outperforming the strongest baseline at that
period by 7.80 points. These results demonstrate that incorporating
the SSL representation provides substantial benefits for temporal
generalization, with particularly consistent gains from Co-Train.

\begin{table*}[t]
\centering
\normalsize
\setlength{\tabcolsep}{2pt}
\caption{Node classification accuracy (\%) on \texttt{OGB-Arxiv}
across temporal splits and GNN backbones.}
\label{tab:arxiv-results}
\begin{adjustbox}{max width=\textwidth}

\begin{tabular*}{\textwidth}{@{\extracolsep{\fill}}l|ccc|l|ccc@{}}
\toprule
\textbf{Method (SAGE)} & 2014--2016 & 2016--2018 & 2018--2020 &
\textbf{Method (GAT)} & 2014--2016 & 2016--2018 & 2018--2020\\
\midrule

ERM
& 42.56 $\pm$ 0.91 & 40.53 $\pm$ 2.03 & 37.19 $\pm$ 2.22
&
ERM
& 43.36 $\pm$ 1.04 & 40.42 $\pm$ 1.89 & 37.84 $\pm$ 1.88\\

EERM
& 41.55 $\pm$ 0.68 & 40.36 $\pm$ 1.29 & 38.95 $\pm$ 1.57
&
EERM
& 49.24 $\pm$ 0.47 & 48.54 $\pm$ 0.66 & 45.53 $\pm$ 0.93\\

LiSA
& 45.68 $\pm$ 2.55 & 43.64 $\pm$ 3.26 & 41.04 $\pm$ 3.00
&
LiSA
& 49.82 $\pm$ 0.13 & 48.34 $\pm$ 0.09 & 44.46 $\pm$ 0.22\\

MARIO
& 32.75 $\pm$ 1.58 & 28.65 $\pm$ 1.81 & 24.08 $\pm$ 1.94
&
MARIO
& 41.08 $\pm$ 1.12 & 38.30 $\pm$ 1.09 & 34.66 $\pm$ 1.15\\

DGI
& {38.62 $\pm$ 0.71}
& {35.90 $\pm$ 1.00}
& {33.80 $\pm$ 1.29}
&
DGI
& {41.34 $\pm$ 0.89}
& {38.13 $\pm$ 1.27}
& {35.77 $\pm$ 1.49}\\

GRACE
& {43.22 $\pm$ 0.39}
& {40.68 $\pm$ 0.72}
& {37.05 $\pm$ 0.64}
&
GRACE
& {44.60 $\pm$ 0.33}
& {42.42 $\pm$ 0.48}
& {38.63 $\pm$ 0.72}\\

CoTrain-DGI
& {49.77 $\pm$ 0.55}
& {48.20 $\pm$ 0.62}
& {45.18 $\pm$ 0.74}
&
CoTrain-DGI
& {50.62 $\pm$ 0.36}
& {48.72 $\pm$ 0.47}
& {44.95 $\pm$ 0.51}\\

CoTrain-GRACE
& {49.71 $\pm$ 0.40}
& {48.22 $\pm$ 0.76}
& {45.22 $\pm$ 0.84}
&
CoTrain-GRACE
& {50.21 $\pm$ 0.32}
& {48.34 $\pm$ 0.49}
& {44.43 $\pm$ 0.43}\\

% Retrieval-DGI
% & {43.80 $\pm$ 0.33}
% & {44.12 $\pm$ 0.51}
% & {42.15 $\pm$ 0.71}
% &
% Retrieval-DGI
% & {48.22 $\pm$ 0.34}
% & {47.61 $\pm$ 0.38}
% & {45.40 $\pm$ 0.48}\\

% Retrieval-GRACE
% & {46.90 $\pm$ 0.96}
% & {49.58 $\pm$ 1.26}
% & {51.69 $\pm$ 1.59}
% &
% Retrieval-GRACE
% & {47.25 $\pm$ 0.98}
% & {46.57 $\pm$ 1.72}
% & {48.68 $\pm$ 2.22}\\
% Retrieval-GRACE
% & {38.97 $\pm$ 0.19}
% & {40.05 $\pm$ 0.15}
% & {37.91 $\pm$ 0.20}
% &
% Retrieval-GRACE
% & {40.63 $\pm$ 0.22}
% & {40.95 $\pm$ 0.66}
% & {38.39 $\pm$ 0.63}\\
% Retrieval\textsuperscript{*}-DGI
% & {50.46 $\pm$ 1.68}
% & {51.70 $\pm$ 1.97}
% & {53.67 $\pm$ 2.04}
% &
% Retrieval\textsuperscript{*}-DGI
% & {56.31 $\pm$ 1.28}
% & {55.18 $\pm$ 1.63}
% & {57.46 $\pm$ 1.78}\\

% Retrieval\textsuperscript{*}-GRACE
% & {47.44 $\pm$ 1.01}
% & {50.38 $\pm$ 1.34}
% & {52.58 $\pm$ 1.47}
% &
% Retrieval\textsuperscript{*}-GRACE
% & {47.50 $\pm$ 1.03}
% & {46.51 $\pm$ 1.84}
% & {48.35 $\pm$ 2.19}\\
Retrieval-DGI
& {43.41 $\pm$ 0.52}
& {43.74 $\pm$ 0.74}
& {41.99 $\pm$ 0.47}
&
Retrieval-DGI
& {48.33 $\pm$ 0.22}
& {47.47 $\pm$ 0.29}
& {45.19 $\pm$ 0.25}\\

Retrieval-GRACE
& {41.98 $\pm$ 0.75}
& {41.92 $\pm$ 0.99}
& {39.97 $\pm$ 1.06}
&
Retrieval-GRACE
& {47.74 $\pm$ 0.18}
& {47.07 $\pm$ 0.25}
& {44.93 $\pm$ 0.29}\\

Retrieval\textsuperscript{*}-DGI
& \cellcolor{tan}{52.01 $\pm$ 1.40}
& \cellcolor{tan}{54.15 $\pm$ 1.89}
& \cellcolor{tan}{56.66 $\pm$ 1.90}
&
Retrieval\textsuperscript{*}-DGI
& \cellcolor{tan}{54.21 $\pm$ 1.19}
& \cellcolor{tan}{52.09 $\pm$ 1.92}
& \cellcolor{brown!20}{53.86 $\pm$ 2.38}\\

Retrieval\textsuperscript{*}-GRACE
& \cellcolor{brown!20}{51.92 $\pm$ 0.69}
& \cellcolor{brown!20}{53.77 $\pm$ 1.10}
& \cellcolor{brown!20}{56.45 $\pm$ 0.96}
&
Retrieval\textsuperscript{*}-GRACE
& \cellcolor{brown!20}{53.53 $\pm$ 0.71}
& \cellcolor{brown!20}{51.75 $\pm$ 2.00}
& \cellcolor{tan}{54.03 $\pm$ 2.22}\\

\bottomrule
\end{tabular*}
\end{adjustbox}
\end{table*}

\para{Temporal Generalization on Static Graphs.}
We next evaluate temporal generalization on \texttt{OGB-Arxiv}, using
papers published before 2011 for training and three future periods,
2014--2016, 2016--2018, and 2018--2020, for evaluation.
Table~\ref{tab:arxiv-results} reports node classification accuracy with
GraphSAGE and GAT as the supervised backbones, while the SSL branches
use GraphSAGE encoders. For Dual-Space Retrieval, we use class-centroid
retrieval to better represent individual classes in this multiclass
setting; this choice over neighbor-based KNN is further discussed in Table~\ref{tab:arxiv-retrieval-val-knn} in Sec.~\ref{sec:app-centroid-retrieval}. 

Co-Train shows strong and consistent temporal generalization with both
DGI and GRACE. With GraphSAGE, CoTrain-DGI improves over LiSA, the
strongest existing baseline, by 4.09, 4.56, and 4.14 percentage points
across the three temporal splits, while CoTrain-GRACE achieves similar
gains of 4.03, 4.58, and 4.18 points. With GAT, both variants also
outperform the strongest baselines across all three splits.

Dual-Space Retrieval remains competitive on \texttt{OGB-Arxiv},
although it is less effective with GraphSAGE than with GAT. Feature
smoothing consistently improves centroid-based retrieval across both
backbones and SSL objectives. With GraphSAGE, \textit{Retrieval*}
achieves up to 52.01\%, 54.15\%, and 56.66\% across the three splits,
exceeding the strongest existing baseline by 6.33, 10.51, and 15.62
percentage points, respectively. With GAT, it reaches up to 54.21\%,
52.09\%, and 54.03\%. Further analysis shows that feature smoothing
increases neighborhood label purity in both representation spaces
(Table~\ref{tab:arxiv-smoothing-purity} Sec.~\ref{sec:app-feature-smoothing}). 

\subsection{Cross-domain Generalization Performance}
We evaluate cross-domain generalization on two social network benchmarks,
\texttt{Twitch} and \texttt{FB-100}, where models are trained on selected
domains and evaluated on unseen ones. These benchmarks capture domain
shifts across online communities and university networks. The results
further support the use of SSL representations as complementary signals
for supervised OOD generalization, with Co-Train showing consistently
strong performance across unseen domains.
\begin{table*}[t]
\centering
\small
\setlength{\tabcolsep}{1.5pt}
\caption{ROC-AUC scores (\%) on \texttt{Twitch} across domain splits
with different GNN backbones.}
\label{tab:twitch-results}
\begin{tabular*}{\textwidth}{@{\extracolsep{\fill}}l|ccccc|ccccc@{}}
\toprule
 \textbf{Method} & \multicolumn{5}{c}{GCN} & \multicolumn{5}{c}{GCNII} \\
\cmidrule(lr){2-6}\cmidrule(lr){7-11}
 & ES & FR & PTBR & RU & TW & ES & FR & PTBR & RU & TW \\
\midrule
ERM & 54.64 $\pm$ 4.43 & 52.32 $\pm$ 1.03 & 49.87 $\pm$ 6.46 & 51.29 $\pm$ 1.58 & 51.59 $\pm$ 3.34 & 63.21 $\pm$ 0.37 & 59.42 $\pm$ 0.28 & 61.19 $\pm$ 0.39 & 54.99 $\pm$ 0.41 & 58.19 $\pm$ 0.26 \\
EERM & 55.63 $\pm$ 4.21 & 53.78 $\pm$ 2.48 & 52.53 $\pm$ 6.74 & 51.87 $\pm$ 1.40 & 52.43 $\pm$ 2.83 & 63.38 $\pm$ 0.88 & 60.32 $\pm$ 0.50 & 61.22 $\pm$ 0.94 & 54.84 $\pm$ 0.65 & 58.12 $\pm$ 0.34 \\
LiSA & 56.51 $\pm$ 4.26 & 52.43 $\pm$ 1.68 & 56.76 $\pm$ 2.86 & 51.77 $\pm$ 1.41 & 52.11 $\pm$ 1.39 & 55.72 $\pm$ 1.34 & 54.95 $\pm$ 2.32 & 56.12 $\pm$ 1.76 & 51.43 $\pm$ 0.79 & 52.47 $\pm$ 0.73 \\
MARIO & {62.05 $\pm$ 0.94} & {58.81 $\pm$ 0.86} & \cellcolor{brown!20}{62.69 $\pm$ 0.79} & {53.52 $\pm$ 0.82} & {53.20 $\pm$ 0.82} & {61.78 $\pm$ 1.08} & {57.88 $\pm$ 0.72} & {62.87 $\pm$ 0.75} & {53.54 $\pm$ 0.64} & {53.61 $\pm$ 0.34} \\
DGI & {54.60 $\pm$ 4.31} & {54.38 $\pm$ 2.98} & {46.73 $\pm$ 4.69} & {50.97 $\pm$ 1.04} & {49.76 $\pm$ 2.18} & {59.65 $\pm$ 3.05} & {59.29 $\pm$ 2.21} & {55.91 $\pm$ 6.59} & {52.37 $\pm$ 1.17} & {51.13 $\pm$ 1.69} \\
GRACE & {58.22 $\pm$ 1.74} & {61.59 $\pm$ 0.38} & {56.80 $\pm$ 2.07} & {52.46 $\pm$ 0.51} & {51.76 $\pm$ 1.32} & {55.65 $\pm$ 2.68} & {60.37 $\pm$ 1.43} & {57.77 $\pm$ 5.80} & {56.05 $\pm$ 0.62} & {53.42 $\pm$ 1.67} \\
CoTrain-DGI & \cellcolor{brown!20}{64.69 $\pm$ 0.72} & \cellcolor{tan}{63.57 $\pm$ 0.29} & \cellcolor{tan}{62.76 $\pm$ 0.41} & \cellcolor{brown!20}{55.76 $\pm$ 0.55} & \cellcolor{brown!20}{56.80 $\pm$ 0.73} & \cellcolor{tan}{66.09 $\pm$ 0.55} & \cellcolor{tan}{65.07 $\pm$ 0.28} & \cellcolor{tan}{64.37 $\pm$ 0.46} & \cellcolor{tan}{56.46 $\pm$ 0.38} & \cellcolor{tan}{59.31 $\pm$ 0.47} \\
CoTrain-GRACE & \cellcolor{tan}{64.80 $\pm$ 0.52} & \cellcolor{brown!20}{63.55 $\pm$ 0.44} & {62.14 $\pm$ 0.46} & \cellcolor{tan}{56.12 $\pm$ 0.40} & \cellcolor{tan}{56.92 $\pm$ 0.85} & \cellcolor{brown!20}{65.19 $\pm$ 1.02} & \cellcolor{brown!20}{64.70 $\pm$ 0.62} & \cellcolor{brown!20}{63.15 $\pm$ 1.56} & \cellcolor{brown!20}{56.22 $\pm$ 0.46} & \cellcolor{brown!20}{58.66 $\pm$ 0.73} \\
% Retrieval-DGI & {63.58 $\pm$ 0.69} & {60.13 $\pm$ 0.89} & \cellcolor{tan}{63.02 $\pm$ 1.22} & {54.30 $\pm$ 0.70} & {55.66 $\pm$ 1.07} & {62.76 $\pm$ 0.95} & {61.63 $\pm$ 0.32} & {62.21 $\pm$ 1.40} & {53.79 $\pm$ 0.85} & {57.91 $\pm$ 0.70} \\
% Retrieval-GRACE & {62.43 $\pm$ 1.28} & {59.21 $\pm$ 0.54} & {60.86 $\pm$ 1.38} & {54.41 $\pm$ 0.68} & {54.21 $\pm$ 0.41} & {61.95 $\pm$ 1.17} & {60.82 $\pm$ 0.76} & {60.43 $\pm$ 1.68} & {53.67 $\pm$ 0.40} & {56.34 $\pm$ 1.04} \\
Retrieval-DGI
& {63.40 $\pm$ 0.69}
& {59.38 $\pm$ 0.57}
& {62.29 $\pm$ 0.71}
& {54.73 $\pm$ 0.83}
& {55.24 $\pm$ 0.60}
&
{62.27 $\pm$ 1.20}
& {61.60 $\pm$ 0.70}
& {62.73 $\pm$ 1.42}
& {53.98 $\pm$ 0.90}
& {57.42 $\pm$ 0.85}
\\

Retrieval-GRACE
& {61.32 $\pm$ 0.97}
& {58.36 $\pm$ 0.58}
& {60.03 $\pm$ 1.55}
& {54.31 $\pm$ 0.65}
& {54.29 $\pm$ 0.69} 
&
{61.35 $\pm$ 0.94}
& {60.23 $\pm$ 0.72}
& {59.86 $\pm$ 1.18}
& {53.75 $\pm$ 1.33}
& {56.10 $\pm$ 0.55} \\
\bottomrule
\end{tabular*}
\end{table*}

\para{Training on a Single Graph.}
On \texttt{Twitch}, we train on the single-domain graph \texttt{DE}
and evaluate generalization to five unseen domains: \texttt{ES},
\texttt{FR}, \texttt{PTBR}, \texttt{RU}, and \texttt{TW}.
We report ROC-AUC in Table~\ref{tab:twitch-results}, using GCN and
GCNII as the supervised backbones and GCN for the SSL encoders.

Co-Train consistently achieves the strongest performance across the
unseen domains. With GCN, CoTrain-DGI and CoTrain-GRACE obtain average
ROC-AUC scores of 60.72\% and 60.71\%, respectively, improving over
the strongest baseline on average (MARIO, 58.05\%) by 2.66 and 2.65
percentage points. The gains remain consistent with GCNII:
CoTrain-DGI and CoTrain-GRACE achieve 62.26\% and 61.58\% on average,
outperforming the strongest baseline (EERM, 59.58\%) by 2.68 and 2.01
points, respectively. For example, on \texttt{FR}, CoTrain-DGI reaches
65.07\% with GCNII, compared with 60.37\% for the strongest baseline.

Dual-Space Retrieval also remains competitive across unseen domains.
With GCN, Retrieval-DGI achieves an average ROC-AUC of 59.01\%,
exceeding the strongest baseline average by 0.95 points. With GCNII,
it achieves 59.60\%, closely matching the strongest baseline average
of 59.58\%. We additionally report results with GAT as the supervised
backbone in Table~\ref{tab:twitch-gat}, Sec.~\ref{sec:additional_experiment_results}
. Co-Train remains consistently strong, while the performance of Dual-Space Retrieval varies more across backbone architectures.

\begin{table}[t]
\centering
\caption{Node classification accuracy (\%) on \texttt{FB-100} across
different training-graph combinations.}
\label{tab:fb100-pb-results}
\normalsize
\setlength{\tabcolsep}{1pt}
\begin{adjustbox}{max width=\columnwidth}
\begin{tabular}{l|ccc}
\toprule
\textbf{Method} & {\footnotesize JHU+CIT+AMH} & {\footnotesize BIN+DUK+PRI} & {\footnotesize WUSTL+BRD+CMU} \\
\midrule
\multicolumn{4}{c}{Penn} \\
\midrule
ERM & 50.78 $\pm$ 0.03 & 50.73 $\pm$ 1.19 & 51.13 $\pm$ 0.83 \\
EERM & 49.58 $\pm$ 1.17 & 50.42 $\pm$ 0.75 & 50.78 $\pm$ 0.60 \\
LiSA & 50.84 $\pm$ 1.24 & 50.92 $\pm$ 0.10 & 50.81 $\pm$ 0.02 \\
MARIO & \cellcolor{brown!20}53.73 $\pm$ 1.22 & \cellcolor{brown!20}53.46 $\pm$ 1.16 & 52.99 $\pm$ 1.18 \\
DGI & {51.48 $\pm$ 2.06} & {51.44 $\pm$ 1.00} & {50.84 $\pm$ 1.13} \\
GRACE & {53.00 $\pm$ 0.82} & {52.25 $\pm$ 0.59} & {52.48 $\pm$ 0.92} \\
CoTrain-DGI & \cellcolor{tan}{53.76 $\pm$ 0.97} & \cellcolor{tan}{54.06 $\pm$ 1.53} & {53.57 $\pm$ 0.74} \\
CoTrain-GRACE & {53.67 $\pm$ 0.72} & {53.11 $\pm$ 1.65} & \cellcolor{brown!20}{54.43 $\pm$ 0.87} \\
Retrieval-DGI
& {53.32 $\pm$ 0.37}
& {53.04 $\pm$ 0.49}
& \cellcolor{tan}{54.69 $\pm$ 0.38} \\

Retrieval-GRACE
& {50.61 $\pm$ 0.41}
& {51.11 $\pm$ 0.36}
& {50.94 $\pm$ 0.53} \\
\midrule
\multicolumn{4}{c}{Brown} \\
\midrule
ERM & 56.82 $\pm$ 0.26 & 51.31 $\pm$ 3.16 & 54.27 $\pm$ 5.77 \\
EERM & 56.82 $\pm$ 0.26 & 51.41 $\pm$ 3.16 & 54.27 $\pm$ 5.77 \\
LiSA & 53.16 $\pm$ 4.84 & 55.67 $\pm$ 1.46 & 54.70 $\pm$ 4.65 \\
MARIO & 50.14 $\pm$ 3.15 & 55.93 $\pm$ 0.89 & 54.78 $\pm$ 3.67 \\
DGI & {46.33 $\pm$ 3.69} & {47.59 $\pm$ 2.93} & {45.71 $\pm$ 2.07} \\
GRACE & {56.00 $\pm$ 0.75} & {54.39 $\pm$ 2.14} & \cellcolor{brown!20}{56.98 $\pm$ 0.53} \\
CoTrain-DGI & \cellcolor{brown!20}{57.14 $\pm$ 1.09} & \cellcolor{brown!20}{56.78 $\pm$ 0.46} & {55.30 $\pm$ 3.40} \\
CoTrain-GRACE & \cellcolor{tan}{57.32 $\pm$ 0.96} & {56.30 $\pm$ 1.86} & {53.92 $\pm$ 2.77} \\
Retrieval-DGI
& {56.53 $\pm$ 0.69}
& \cellcolor{tan}{57.50 $\pm$ 0.52}
& \cellcolor{tan}{57.19 $\pm$ 0.47} \\
Retrieval-GRACE
& {51.92 $\pm$ 0.69}
& {52.39 $\pm$ 1.17}
& {50.96 $\pm$ 1.24} \\
\bottomrule
\end{tabular}
\end{adjustbox}

\end{table}

\para{Training on Multiple Graphs.}
On \texttt{FB-100}, we train on graphs from three universities and
evaluate on unseen university networks. We use GCN as the supervised
backbone and GCN encoders for DGI and GRACE, and report test accuracy
in Table~\ref{tab:fb100-pb-results}.

Both Co-Train and Dual-Space Retrieval achieve strong performance on
Penn and Brown. On Penn, Retrieval-DGI reaches 54.69\% under the WUSTL+BRD+CMU source combination, outperforming the strongest baseline
(52.99\%) by 1.70 percentage points. Co-Train also consistently
improves or matches the strongest baselines across the source
combinations. Similar gains are observed on Brown: Retrieval-DGI
achieves 57.50\% and 57.19\% under BIN+DUK+PRI and WUSTL+BRD+CMU, exceeding the
strongest baselines by 1.57 and 2.41 points, respectively, while
Co-Train remains consistently competitive.

Additional results on Texas are reported in Table~\ref{tab:app-fb100-texas}, Sec.~\ref{sec:additional_experiment_results}. While the improvements are more modest, Co-Train and Retrieval remain competitive, with Co-Train-DGI achieving the best performance under  WUSTL+BRD+CMU (56.89\%).

\subsection{Ablation Study}
\label{sec:ablation}
We examine whether self-supervised representations provide complementary information to supervised representations. For both Co-Train and Dual-Space Retrieval, we compare the full method with its supervised-only and SSL-only variants, using DGI and GRACE as the SSL objectives.

\begin{figure}[t]
    \centering

    \begin{subfigure}[t]{0.48\columnwidth}
        \centering
        \includegraphics[
            width=\linewidth,
            trim={0 0 570.8 0},
            clip
        ]{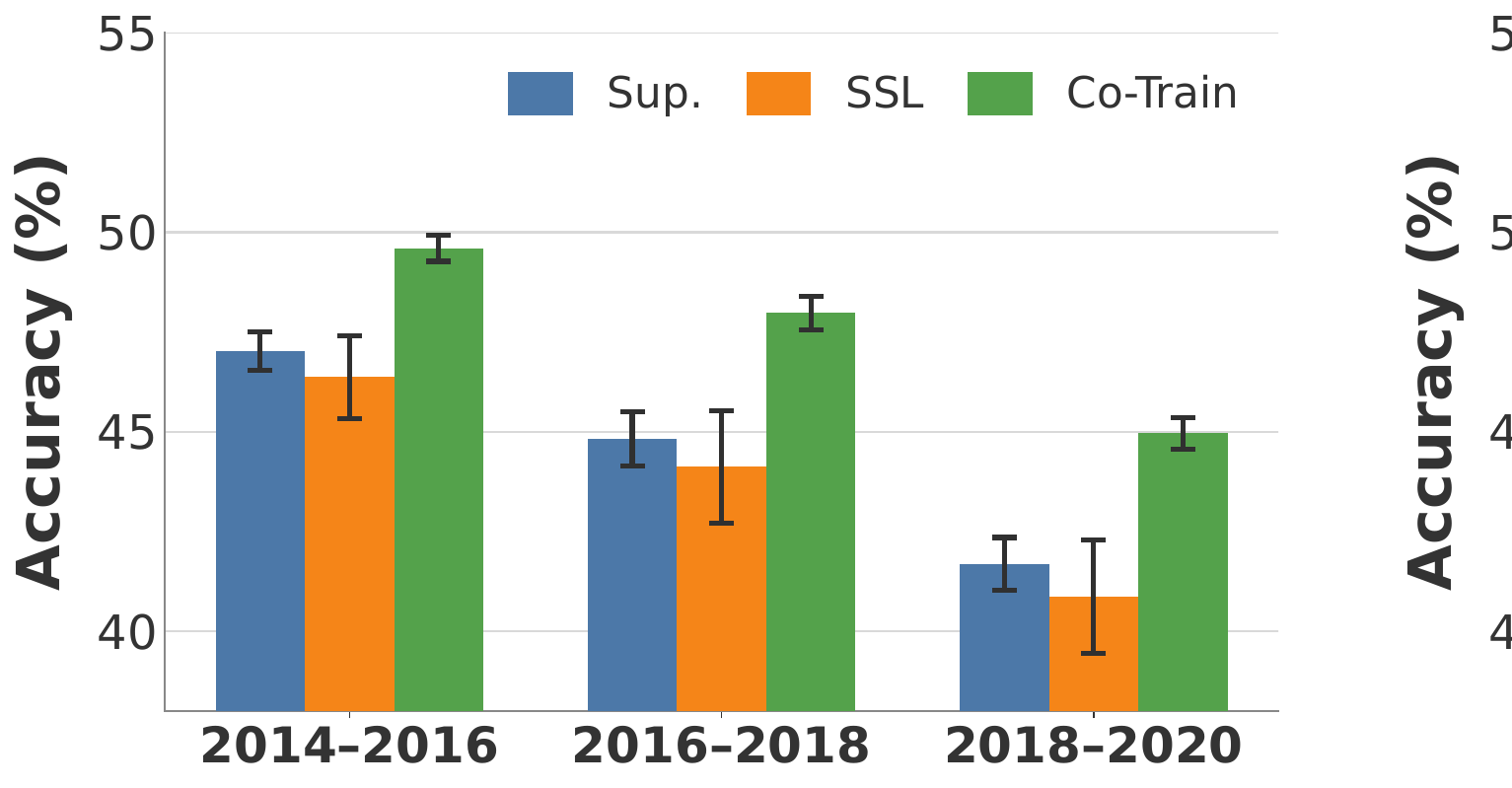}
        \caption{DGI}
        \label{fig:arxiv_sage_cotrain_dgi}
    \end{subfigure}
    \hfill
    \begin{subfigure}[t]{0.48\columnwidth}
        \centering
        \includegraphics[
            width=\linewidth,
            trim={570.8 0 0 0},
            clip
        ]{figs/arxiv_sage_cotrain.pdf}
        \caption{GRACE}
        \label{fig:arxiv_sage_cotrain_grace}
    \end{subfigure}

    \caption{Accuracy comparison of supervised (Sup.), self-supervised
(SSL), and Co-Train models on \texttt{OGB-Arxiv} across three temporal test periods.}
    \label{fig:arxiv_sage_cotrain_ablation}
\end{figure}

For Co-Train, Figure~\ref{fig:arxiv_sage_cotrain_ablation} reports the results on \texttt{OGB-Arxiv} with GraphSAGE as the supervised backbone. Combining supervised and SSL learning consistently outperforms either individual objective across all three temporal splits. With DGI, Co-Train improves over the stronger individual branch by 2.58, 3.16, and 3.27 percentage points on the three periods (2014--2016, 2016--2018, and 2018--2020), respectively. With GRACE, Co-Train similarly outperforms the stronger individual branch by 2.40, 2.50, and 2.96 points, respectively. These results demonstrate the benefit of incorporating complementary SSL signals during representation learning.

\begin{figure}[t]
    \centering

    \begin{subfigure}[t]{0.48\columnwidth}
        \centering
        \includegraphics[
            width=\linewidth,
            trim={0 0 570.7 0},
            clip
        ]{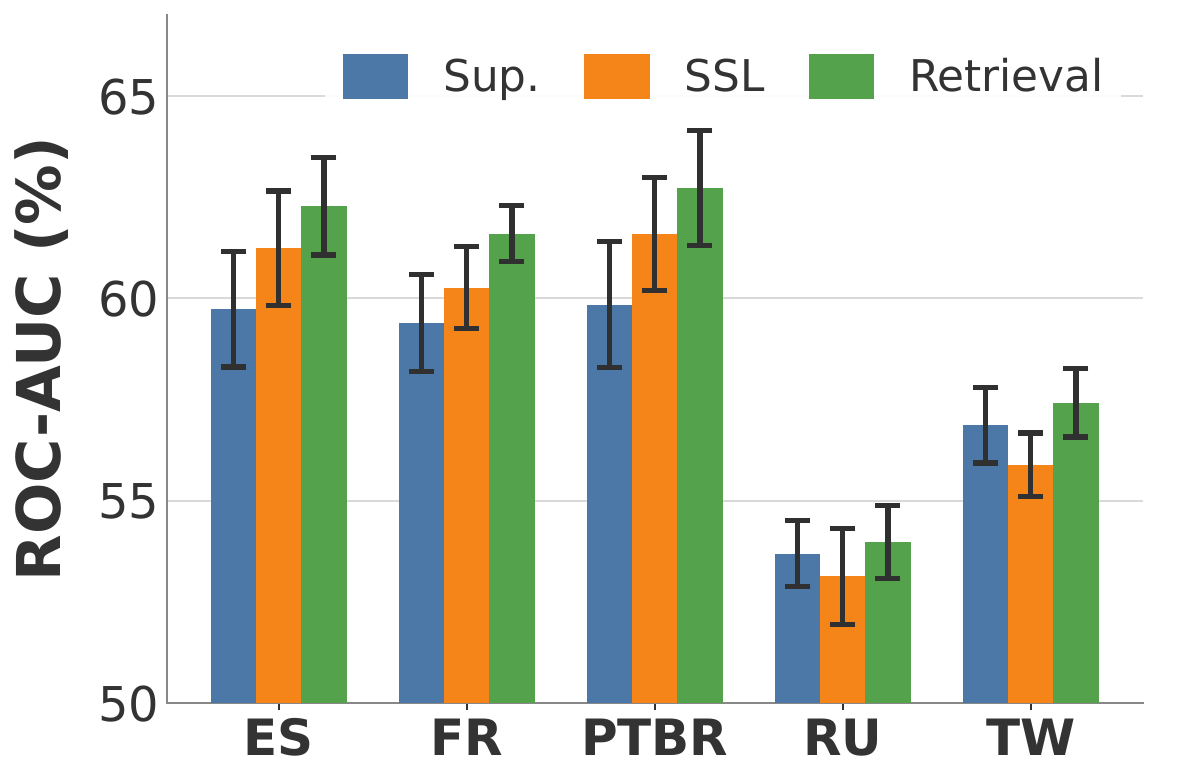}
        \caption{DGI}
        \label{fig:twitch_gcnii_retrieval_dgi}
    \end{subfigure}
    \hfill
    \begin{subfigure}[t]{0.48\columnwidth}
        \centering
        \includegraphics[
            width=\linewidth,
            trim={570.7 0 0 0},
            clip
        ]{figs/twitch_gcnii_retrieval.pdf}
        \caption{GRACE}
        \label{fig:twitch_gcnii_retrieval_grace}
    \end{subfigure}

    \caption{ROC-AUC comparsion of supervised (Sup.), self-supervised (SSL),
    and Retrieval models on five \texttt{Twitch} domains.}
    \label{fig:twitch_gcnii_retrieval_ablation}
\end{figure}

For Dual-Space Retrieval, Figure~\ref{fig:twitch_gcnii_retrieval_ablation} reports the results on \texttt{Twitch} with GCNII as the supervised backbone. With DGI, combining the two retrieval spaces outperforms either individual space across all five domains, with gains of up to 2.89 percentage points. With GRACE, the combined prediction improves over both individual spaces on ES, FR, and PTBR, while remaining comparable on RU and TW. These results show that supervised and SSL representations can provide complementary retrieval signals at inference time.

We then compare gated fusion in Co-Train with additive and
concatenation-based fusion. Using \texttt{Twitch} with GCNII as the
supervised backbone, gated fusion achieves a slightly higher average
ROC-AUC of 62.16\%, compared with 62.04\% for addition and 61.95\%
for concatenation as shown in Table~\ref{tab:twitch_fusion_ablation}. Overall,
the three fusion strategies perform comparably. We use gated fusion
because it allows adaptive weighting of the supervised and SSL
representations.

% \begin{figure}[t]
%     \centering

%     \includegraphics[width=\columnwidth]
%     {figs/arxiv_sage_cotrain.pdf}
%     \caption{Accuracy comparison of supervised (Sup.), self-supervised (SSL),
%     and combined co-training models on the temporally shifted Arxiv-SAGE
%     test windows T1--T3.}
%     \label{fig:arxiv_sage_cotrain_abaltion}
% \end{figure}
% \begin{figure}[t]
%     \centering
%     \includegraphics[width=\columnwidth]{figs/twitch_gcnii_retrieval.pdf}
%     \caption{Retrieval performance measured by ROC-AUC on five Twitch domains. }
%     \label{fig:twitch_gcnii_retrieval_abaltion}
% \end{figure}
% \begin{figure}[t]
%     \centering
%     \includegraphics[width=\columnwidth]{figs/separation.pdf}
%     \caption{Comparison of representation-separation scores for supervised,
%     self-supervised, and co-trained models under the DGI and GRACE objectives on Arxiv-SAGE.}
%     \label{fig:separation}
% \end{figure}
\begin{figure}[t]
    \centering

    \begin{subfigure}[t]{0.48\columnwidth}
        \centering
        \includegraphics[
            width=\linewidth,
            trim={0 0 527.8 0},
            clip
        ]{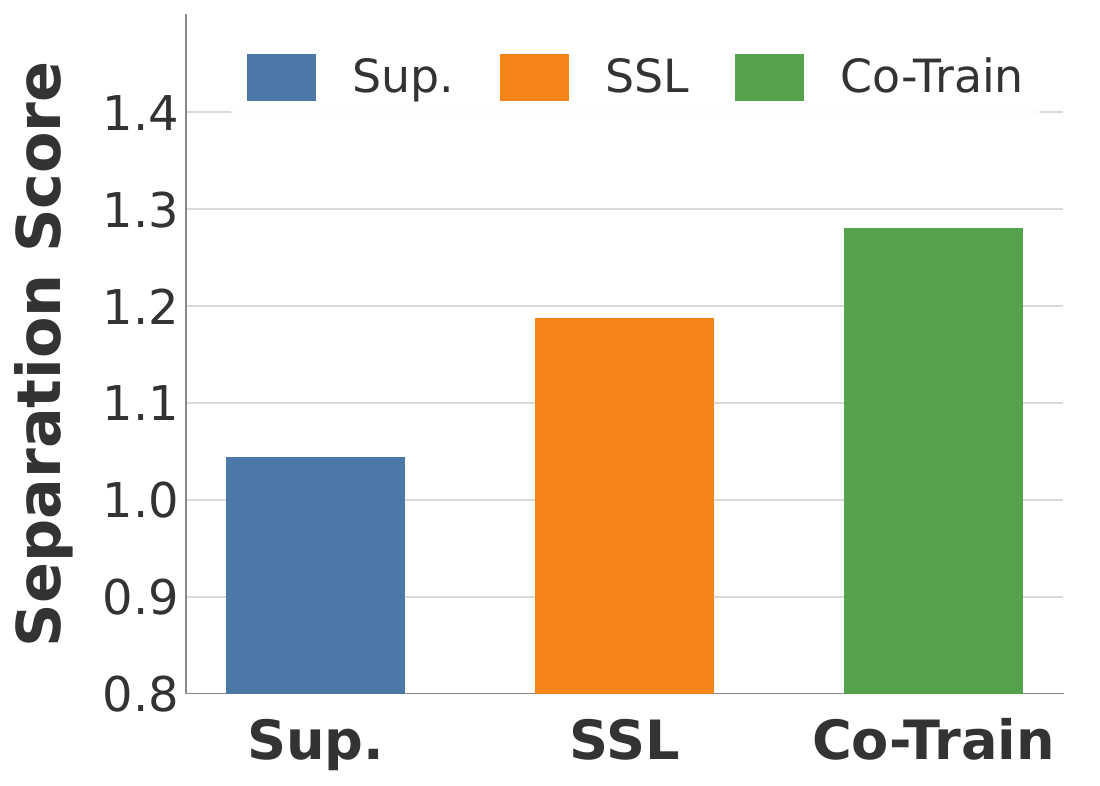}
        \caption{DGI}
        \label{fig:separation_dgi}
    \end{subfigure}
    \hfill
    \begin{subfigure}[t]{0.48\columnwidth}
        \centering
        \includegraphics[
            width=\linewidth,
            trim={527.8 0 0 0},
            clip
        ]{figs/separation.pdf}
        \caption{GRACE}
        \label{fig:separation_grace}
    \end{subfigure}

    \caption{Comparison of representation-separation scores for supervised,
self-supervised (SSL), and Co-Train models with on
\texttt{OGB-Arxiv}.}
    \label{fig:separation}
\end{figure}

\begin{figure}[h]
    \centering
    \begin{subfigure}[t]{0.32\linewidth}
        \centering
        \includegraphics[
            width=\linewidth,
            trim=0pt 0pt 520pt 0pt,
            clip
        ]{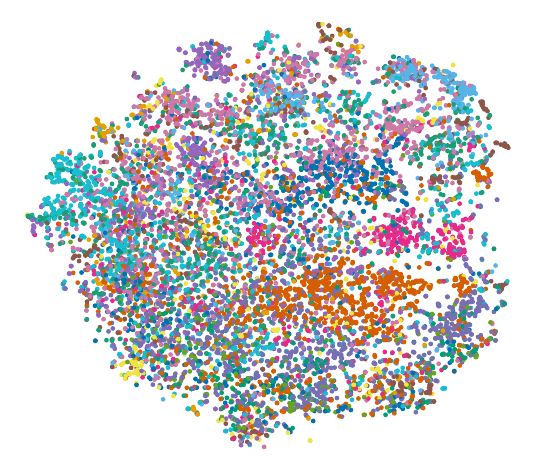}
        \caption{Supervised}
    \end{subfigure}
    \hfill
    \begin{subfigure}[t]{0.32\linewidth}
        \centering
        \includegraphics[
            width=\linewidth,
            trim=260pt 0pt 260pt 0pt,
            clip
        ]{figs/grace_15classes.pdf}
        \caption{Self-Supervised}
    \end{subfigure}
    \hfill
    \begin{subfigure}[t]{0.32\linewidth}
        \centering
        \includegraphics[
            width=\linewidth,
            trim=520pt 0pt 0pt 0pt,
            clip
        ]{figs/grace_15classes.pdf}
        \caption{Co-Train (GRACE)}
    \end{subfigure}
    \captionsetup{name={Figure}}
    \caption{t-SNE visualization of supervised, self-supervised (SSL), and
Co-Train test-node representations on the 2018--2020 temporal split of
\texttt{OGB-Arxiv} with GRACE.}
    \label{fig:grace-tsne}
\end{figure}

\begin{table}[ht]
    \centering
    \normalsize
    \setlength{\tabcolsep}{3pt}
    \caption{ROC-AUC (\%) of different Co-Train fusion operators across
\texttt{Twitch} domains with GCNII.}
    \label{tab:twitch_fusion_ablation}

    \begin{tabular}{l|ccc}
        \toprule
        \textbf{Domain} & Gate &Add. &Concat. \\
        \midrule
        ES   & 65.88 $\pm$ 0.66 & \textbf{65.88 $\pm$ 0.26} & 65.77 $\pm$ 0.64 \\
        FR   & \textbf{64.89 $\pm$ 0.36} & 64.62 $\pm$ 0.49 & 64.64 $\pm$ 0.46 \\
        PTBR & \textbf{64.40 $\pm$ 0.63} & 64.26 $\pm$ 0.86 & 64.03 $\pm$ 0.70 \\
        RU   & \textbf{56.57 $\pm$ 0.29} & 56.43 $\pm$ 0.41 & 56.49 $\pm$ 0.49 \\
        TW   & \textbf{59.07 $\pm$ 0.52} & 59.03 $\pm$ 0.71 & 58.82 $\pm$ 0.57 \\
        \midrule
        Avg. & \textbf{62.16} & 62.04 & 61.95 \\
        \bottomrule
    \end{tabular}

\end{table}

\subsection{Additional Analysis}

\para{Representation Separation.}
We further examine how Co-Train affects the learned representation
space on \texttt{OGB-Arxiv}. Figure~\ref{fig:grace-tsne} visualizes
test-node representations from 15 randomly selected classes in the final
temporal split (2018--2020) using t-SNE, with GraphSAGE as both the
supervised backbone and SSL encoder and GRACE as the SSL objective.
Compared with either representation alone, Co-Train produces more
compact class structures with clearer inter-class boundaries, indicating
a more discriminative representation space. A corresponding
visualization with DGI exhibits a similar pattern
(Fig.~\ref{fig:tsne-dgi-appendix}, Sec.~\ref{sec:additional_results}).
We further quantify this effect using the separation score in
Figure~\ref{fig:separation}, defined as the ratio of inter-class to
intra-class distance. We first L2-normalize the embeddings and then
compute the inter- and intra-class Euclidean distances. A larger score
indicates better class separation. Co-Train achieves the highest
separation score with both DGI and GRACE, further showing that
combining supervised and SSL learning leads to better separated
representations.

\para{Efficiency.}
We compare the runtime and memory usage of different methods on
\texttt{OGB-Arxiv} using a 64GB GPU. For a fair comparison, GraphSAGE
is used as the supervised backbone for our methods and as the backbone
for all compared methods; our SSL branch also uses a GraphSAGE encoder.
Each method is evaluated under the configuration used to obtain its best
reported performance. Co-Train is the fastest method in our comparison
while maintaining strong OOD performance, and centroid-based Dual-Space
Retrieval achieves runtime comparable to ERM. All methods fit within GPU
memory. Detailed results are reported in Table~\ref{tab:runtime_memory}, Sec.~\ref{sec:app-run}. 

\section{Conclusion}
\label{sec:conclusion}
In this work, we study how self-supervised representations can
complement supervised learning for graph OOD generalization. We
propose Co-Train, which jointly learns supervised and self-supervised
representations, and Dual-Space Retrieval, which combines predictions
from the two representation spaces at inference time. Across multiple
graph datasets and distribution shifts, both methods achieve strong
performance, with Co-Train consistently outperforming strong OOD
generalization baselines.

Our ablation and representation analyses show that SSL provides
complementary information to supervised learning and improves
representation separation. Overall, our results show that incorporating
SSL as a complementary signal can improve supervised graph OOD
generalization.
\bibliographystyle{ACM-Reference-Format}
\bibliography{bibfile}

@string{naacl = "Proc. of {NAACL}"}

@string{www = "Proc. of {WWW}"}

@string{kdd = "Proc. of {KDD}"}

@string{ccs = "Proc. of {CCS}"}

@string{icml-grl = "Proc. of {ICML} Workshop on Graph Representation Learning and Beyond"}

@string{icml="Proc. of {ICML}"}

@string{cvpr="Proc. of {CVPR}"}

@string{iccv="Proc. of {ICCV}"}

@string{aaai="Proc. of {AAAI}"}

@string{iclr = "Proc. of {ICLR}"}

@string{nips="Proc. of {NeurIPS}"}

@string{wsdm = "Proc. of {WSDM}"}

@inproceedings{wu2022eerm,
title = {Handling Distribution Shifts on Graphs: An Invariance Perspective},
author = {Qitian Wu and Hengrui Zhang and Junchi Yan and David Wipf},
booktitle =iclr,
year = {2022}
}

@inproceedings{sun2019infograph,
  title={InfoGraph: Unsupervised and Semi-supervised Graph-Level Representation Learning via Mutual Information Maximization},
  author={Sun, Fan-Yun and Hoffmann, Jordan and Verma, Vikas and Tang, Jian},
  booktitle=iclr,
  year={2019}
}

@inproceedings{andyZhouKG,
author = {Zhou, Andy and Xu, Xiaojun and Raghunathan, Ramesh and Lal, Alok and Guan, Xinze and Yu, Bin and Li, Bo},
title = {KnowGraph: Knowledge-Enabled Anomaly Detection via Logical Reasoning on Graph Data},
year = {2024},
booktitle = ccs
}

@article{twitch,
author = {Rozemberczki, Benedek and Allen, Carl and Sarkar, Rik},
year = {2021},
title = {Multi-Scale attributed node embedding},
journal = {Journal of Complex Networks},
}

@inproceedings{elliptic,
author = {Aldo Pareja and Giacomo Domeniconi and Jie Chen and
Tengfei Ma and Toyotaro Suzumura and Hiroki Kanezashi and Tim Kaler and Tao B. Schardl and Charles E. Leiserson},
title = {EvolveGCN: Evolving Graph Convolutional Networks for Dynamic Graphs},
booktitle = aaai,
year = {2020}
}

@inproceedings{ogb-arxiv,
 author = {Hu, Weihua and Fey, Matthias and Zitnik, Marinka and Dong, Yuxiao and Ren, Hongyu and Liu, Bowen and Catasta, Michele and Leskovec, Jure},
 booktitle = nips,
title = {Open Graph Benchmark: Datasets for Machine Learning on Graphs},
year = {2020}
}

@inproceedings{lisa,
author={Junchi Yu and Jian Liang and Ran He},
title={Mind the Label Shift of Augmentation-based Graph OOD Generalization}, 
booktitle=cvpr,
year ={2023}
}

@inproceedings{dgi,
title={Deep Graph Infomax}, 
author={Petar Veličković and William Fedus and William L. Hamilton and Pietro Liò and Yoshua Bengio and R Devon Hjelm},
year={2019},
booktitle=iclr
}

@inproceedings{adanpc,
author = {Zhang, Yi-Fan and Wang, Xue and Jin, Kexin and Yuan, Kun and Zhang, Zhang and Wang, Liang and Jin, Rong and Tan, Tieniu},
title = {AdaNPC: exploring non-parametric classifier for test-time adaptation},
year = {2023},
booktitle =icml
}

@inproceedings{mario_unsupervised,
author = {Zhu, Yun and Shi, Haizhou and Zhang, Zhenshuo and Tang, Siliang},
title = {MARIO: Model Agnostic Recipe for Improving OOD Generalization of Graph Contrastive Learning},
year = {2024},
booktitle = www,
}

@inproceedings{goodd_un,
  title={GOOD-D: On Unsupervised Graph Out-Of-Distribution Detection},
  author={Liu, Yixin and Ding, Kaize and Liu, Huan and Pan, Shirui},
  booktitle=wsdm,
  year={2023}
}

@inproceedings{sun2022knnood,
  title={Out-of-distribution Detection with Deep Nearest Neighbors},
  author={Sun, Yiyou and Ming, Yifei and Zhu, Xiaojin and Li, Yixuan},
  booktitle=icml,
  year={2022}
}

@inproceedings{park2023nearest,
  title={Nearest Neighbor Guidance for Out-of-Distribution Detection},
  author={Park, Jaewoo and Jung, Yoon Gyo and Teoh, Andrew Beng Jin},
  booktitle=iccv,
  year={2023}
}

@inproceedings{erm,
 author = {Vapnik, V.},
 booktitle = nips,
 title = {Principles of Risk Minimization for Learning Theory},
 year = {1991}
}

@inproceedings{wu2022dir,
title={Discovering Invariant Rationales for Graph Neural Networks},
author={Ying-Xin Wu and Xiang Wang and An Zhang and Xiangnan He and Tat-seng Chua},
booktitle=iclr,
year={2022},
}

@inproceedings{wu2023gnnsafe,
title = {Energy-based Out-of-Distribution Detection for Graph Neural Networks},
author = {Qitian Wu and Yiting Chen and Chenxiao Yang and Junchi Yan},
booktitle = iclr,
year = {2023}
}

@inproceedings{suresh2021adversarial,
  title={Adversarial Graph Augmentation to Improve Graph Contrastive Learning},
  author={Suresh, Susheel and Li, Pan and Hao, Cong and Neville, Jennifer},
  booktitle=nips,
  year={2021}
}

@inproceedings{gcn,
title={Semi-Supervised Classification with Graph Convolutional Networks},
author={Kipf, Thomas N and Welling, Max},
booktitle=iclr,
year={2017}
}

@inproceedings{graphsage,
author = {Hamilton, William L. and Ying, Rex and Leskovec, Jure},
title = {Inductive Representation Learning on Large Graphs},
booktitle = nips,
year = {2017}
}

@InProceedings{gcn2,
  title = {Simple and Deep Graph Convolutional Networks},
  author = {Chen, Ming and Wei, Zhewei and Huang, Zengfeng and Ding, Bolin and Li, Yaliang},
  booktitle = icml,
  year = 	 {2020},
}

@inproceedings{gpr,
title={Adaptive Universal Generalized PageRank Graph Neural Network},
author={Eli Chien and Jianhao Peng and Pan Li and Olgica Milenkovic},
booktitle=iclr,
year={2021},
}

@inproceedings{gat,
title={Graph Attention Networks},
author={Veličković, Petar and Cucurull, Guillem and Casanova, Arantxa and Romero, Adriana and Liò, Pietro and Bengio, Yoshua},
booktitle=iclr,
year={2018}
}

@inproceedings{graphsaint,
  title={GraphSAINT: Graph Sampling Based Inductive Learning Method},
  author={Hanqing Zeng and Hongkuan Zhou and Ajitesh Srivastava and Rajgopal Kannan and Viktor Prasanna},
  booktitle=iclr,
  year={2020}
}

@inproceedings{defferrard2016convolutional,
title={Convolutional neural networks on graphs with fast localized spectral filtering},
author={Defferrard, Micha{\"e}l and Bresson, Xavier and Vandergheynst, Pierre},
booktitle= nips,
year={2016}
}

@article{twitch_dataset,
    author = {Rozemberczki, Benedek and Allen, Carl and Sarkar, Rik},
    title = {Multi-Scale attributed node embedding},
    journal = {Journal of Complex Networks},
    year = {2021},
}

@article{fb100_dataset,
author = {Traud, Amanda L. and Mucha, Peter J. and Porter, Mason A.},
year = {2012},
title = {Social Structure of Facebook Networks},
journal = {Physica A: Statistical Mechanics and its Applications},
}

@inproceedings{elliptic_dataset,
  title     = {{EvolveGCN: Evolving Graph Convolutional Networks for Dynamic Graphs}},
  author    = {Aldo Pareja and Giacomo Domeniconi and Jie Chen and Tengfei Ma and Toyotaro Suzumura and Hiroki Kanezashi and Tim Kaler and Tao B. Schardl and Charles E. Leiserson},
  booktitle = aaai,
  year      = {2020},
}

@article{papernot2018deep,
  title     = {Deep k-Nearest Neighbors: Towards Confident, Interpretable and Robust Deep Learning},
  author    = {Papernot, Nicolas and McDaniel, Patrick},
  journal   = {arXiv preprint arXiv:1803.04765},
  year      = {2018}
}

@inproceedings{yilmaz-toraman-2022-d2u,
title = "{D}2{U}: Distance-to-Uniform Learning for Out-of-Scope Detection",
author = "Yilmaz, Eyup  and Toraman, Cagri",
booktitle = naacl,
year = {2022}
}

@article{real-world-problem1,
  title        = {Towards Out-Of-Distribution Generalization: A Survey},
  author       = {Shen, Zheyan and Liu, Jiashuo and He, Yue and Zhang, Xingxuan and Xu, Renzhe and Yu, Han and Cui, Peng},
  year         = {2021},
  journal       = {arXiv preprint arXiv:2108.13624},
}

@inproceedings{zhang2022dynamic,
  title     = {Dynamic Graph Neural Networks Under Spatio-Temporal Distribution Shift},
  author    = {Zhang, Zeyang and Wang, Xin and Zhang, Ziwei and Li, Haoyang and Qin, Zhou and Zhu, Wenwu},
  booktitle = nips,
  year      = {2022}
}

@inproceedings{ying2018pinsage,
  title     = {Graph Convolutional Neural Networks for Web‑Scale Recommender Systems},
  author    = {Ying, Rex and He, Ruining and Chen, Kaifeng and Eksombatchai, Pong and Hamilton, William L. and Leskovec, Jure},
  booktitle = kdd,
  year      = {2018}
}

@inproceedings{goel2019end,
  title={An End-to-End Network for Generating Social Relationship Graphs},
  author={Goel, Arushi and Ma, Keng Teck and Tan, Cheston},
  booktitle=cvpr,
  year={2019}
}

@article{buterez2024transfer,
  title   = {Transfer Learning with Graph Neural Networks for Improved Molecular Property Prediction in the Multi‑Fidelity Setting},
  author  = {Buterez, David and Janet, Jon Paul and Kiddle, Steven J. and Oglic, Dino and Li{\`o}, Pietro},
  journal = {Nature Communications},
  year    = {2024},
}

@inproceedings{zhu2021shift,
  title     = {Shift‑Robust GNNs: Overcoming the Limitations of Localized Graph Training Data},
  author    = {Zhu, Qi and Ponomareva, Natalia and Han, Jiawei and Perozzi, Bryan},
  booktitle = nips,
  year      = {2021}
}

@InProceedings{chen2022ciga,
  title       = {Learning Causally Invariant Representations for Out-of-Distribution Generalization on Graphs},
  author      = {Yongqiang Chen and Yonggang Zhang and Yatao Bian and Han Yang and Kaili Ma and Binghui Xie and Tongliang Liu and Bo Han and James Cheng},
  booktitle   = nips,
  year        = {2022}
}

@inproceedings{wu2024canet,
      title = {Graph Out-of-Distribution Generalization via Causal Intervention},
      author = {Qitian Wu and Nie Fan and Chenxiao Yang and Tianyi Bao and Junchi Yan},
      booktitle = www,
      year = {2024}
}

@article{trivedi2021augmentations,
  title={Augmentations in Graph Contrastive Learning: Current Methodological Flaws \& Towards Better Practices},
  author={Trivedi, Puja and Lubana, Ekdeep Singh and Yan, Yujun and Yang, Yaoqing and Koutra, Danai},
  journal={arXiv preprint arXiv:2111.03220},
  year={2021},
}

@inproceedings{shen2024beyond,
  title={Beyond Redundancy: Information-aware Unsupervised Multiplex Graph Structure Learning},
  author={Shen, Zhixiang and Wang, Shuo and Kang, Zhao},
  booktitle= nips,
  year={2024}
}

@inproceedings{liu2022towards,
  title={Towards Unsupervised Deep Graph Structure Learning},
  author={Liu, Yixin and Zheng, Yu and Zhang, Daokun and Chen, Hongxu and Peng, Hao and Pan, Shirui},
  booktitle=www,
  year={2022}
}

@ARTICLE{fanGraphCorrelation,
  author={Fan, Shaohua and Wang, Xiao and Shi, Chuan and Cui, Peng and Wang, Bai},
  journal={IEEE Transactions on Pattern Analysis and Machine Intelligence}, 
  title={Generalizing Graph Neural Networks on Out-of-Distribution Graphs}, 
  year={2024},
}

@inproceedings{zhu2020grace,
  title={Deep Graph Contrastive Representation Learning},
  author={Zhu, Yanqiao and Xu, Yichen and Yu, Feng and Liu, Qiang and Wu, Shu and Wang, Liang},
  booktitle=icml-grl,
  year={2020}
}

@article{wang2025generative,
  title={Generative Risk Minimization for Out-of-Distribution Generalization on Graphs},
  author={Wang, Song and Tan, Zhen and Zhu, Yaochen and Zhang, Chuxu and Li, Jundong},
  journal={TMLR},
  year={2025}
}

@inproceedings{ICLR2025_Decoupled,
 author = {Chen, Yuhan and Luo, Yihong and Song, Yifan and Dai, Pengwen and Tang, Jing and Cao, Xiaochun},
 booktitle = iclr,
 title = {Decoupled Graph Energy-based Model for Node Out-of-Distribution Detection on Heterophilic Graphs},
 year = {2025}
}

@inproceedings{qiao2025icml-gcal,
  title     = {{GCAL: Adapting Graph Models to Evolving Domain Shifts}},
  author    = {Qiao, Ziyue and Cai, Qianyi and Dong, Hao and Gu, Jiawei and Wang, Pengyang and Xiao, Meng and Luo, Xiao and Xiong, Hui},
  booktitle = icml,
  year      = {2025}
}
% \clearpage
\appendix
\definecolor{lightblue}{rgb}{0.8, 0.88, 1}
\definecolor{midblue}{rgb}{0.4, 0.6, 1}
\definecolor{darkblue}{rgb}{0.1, 0.1, 0.8}

\definecolor{lightred}{rgb}{1, 0.8, 0.8}
\definecolor{midred}{rgb}{1, 0.4, 0.4}
\definecolor{darkred}{rgb}{0.8, 0.0, 0.0}
\definecolor{tan}{rgb}{0.82, 0.71, 0.55}

\section{Datasets}
\label{sec:app-data}

Table~\ref{tab:datasets} summarizes the statistics, distribution shifts,
and evaluation metrics of the datasets used in our experiments.

\textbf{Elliptic.}
The Elliptic dataset records Bitcoin transactions across 49 time steps.
Each node represents a transaction, and each edge indicates a payment
flow. Around 20\% of transactions are labeled as licit or illicit, and
the task is to identify illicit transactions in future time steps.
We use time steps 7--11 for training, 12--16 for validation, and
17--49 for testing. The 33 test snapshots are grouped into nine
chronological test sets to evaluate temporal generalization.

\textbf{OGB-Arxiv.}
OGB-Arxiv contains 169,343 Arxiv CS papers from 40 subject areas,
along with their citation links. Both research topics and citation
patterns change over time, resulting in temporal distribution shifts.
The task is to predict a paper's subject area. We use papers published
up to 2011 for training, papers published after 2011 and up to 2014
for validation, and three subsequent periods, 2014--2016, 2016--2018,
and 2018--2020, for testing.

\textbf{Twitch-Explicit (Twitch).}
The Twitch-Explicit dataset contains seven networks, where each node
represents a Twitch user and edges denote mutual friendships. Each
network corresponds to a geographic region, resulting in cross-domain
distribution shifts. The task is to predict whether a Twitch streamer
uses explicit language. We use DE for training, ENGB for validation,
and ES, FR, PTBR, RU, and TW for testing.

\textbf{Facebook-100 (FB100).}
The Facebook-100 dataset comprises 100 Facebook friendship networks
collected in 2005, each corresponding to a different American
university. Nodes represent users, and edges indicate Facebook
friendships. The task is to predict the gender of each user. We use
the selected training universities for training, Cornell and Yale for
validation, and disjoint universities for testing.

\begin{table*}[t]
\centering
\caption{Dataset statistics, distribution shifts, and evaluation metrics.}
  \begin{tabular}{l|ccccc}
    \toprule
    \textbf{Dataset}&Distribution Shift&Nodes & Edges & Classes & Metric \\
    \midrule
    Elliptic \cite{elliptic_dataset} & \cellcolor{midred}Temporal Evolution  &  203,769 & 234,355 & 2 & F1-Score \\
    OGB-Arxiv \cite{hu2020ogb} & \cellcolor{midred}Temporal Evolution & 169,343 & 1,166,243 & 40 & Accuracy \\
    Twitch-explicit \cite{twitch_dataset} & \cellcolor{lightred}Cross-Domain Transfers & 1,912 - 9,498 & 31,299 - 153,138 & 2 & ROC-AUC \\
    Facebook-100 \cite{fb100_dataset}& \cellcolor{lightred}Cross-Domain Transfers& 769 - 41,536 & 16,656 - 1,590,655 & 2 & Accuracy\\
  \bottomrule
\end{tabular}
\label{tab:datasets}
\end{table*}

\begin{table*}[t]
\centering
\small
\setlength{\tabcolsep}{1.5pt}
\caption{ROC-AUC scores (\%) on \texttt{Twitch-Explicit} across domain splits with the GAT backbone.}
\label{tab:twitch-gat}

\begin{tabular}{l|ccccc}
\toprule
\textbf{Method} & ES & FR & PTBR & RU & TW \\
\midrule
ERM & 62.28 $\pm$ 0.88 & 60.06 $\pm$ 1.02 & 60.30 $\pm$ 1.20 & 55.00 $\pm$ 0.40 & \cellcolor{tan}57.55 $\pm$ 0.51 \\
EERM & 63.64 $\pm$ 0.74 & \cellcolor{brown!20}62.30 $\pm$ 0.63 & 61.33 $\pm$ 1.28 & 55.51 $\pm$ 0.54 & 55.77 $\pm$ 0.38 \\
LiSA & 62.96 $\pm$ 0.41 & 61.44 $\pm$ 0.68 & 61.58 $\pm$ 1.01 & 55.25 $\pm$ 0.60 & 56.71 $\pm$ 1.45 \\
MARIO & {61.11 $\pm$ 1.20} & {59.79 $\pm$ 1.90} & {61.12 $\pm$ 0.91} & {54.77 $\pm$ 0.23} & {54.58 $\pm$ 1.34} \\
DGI & {61.19 $\pm$ 1.57} & {57.77 $\pm$ 0.85} & {56.59 $\pm$ 2.79} & {52.96 $\pm$ 0.63} & {54.70 $\pm$ 1.13} \\
GRACE & {63.36 $\pm$ 1.01} & {59.40 $\pm$ 1.20} & {63.95 $\pm$ 0.84} & {53.46 $\pm$ 0.74} & {54.39 $\pm$ 0.18} \\
CoTrain-DGI & \cellcolor{brown!20}{65.03 $\pm$ 0.48} & \cellcolor{tan}{62.36 $\pm$ 0.60} & \cellcolor{brown!20}{63.98 $\pm$ 0.93} & \cellcolor{brown!20}{56.23 $\pm$ 0.63} & {56.88 $\pm$ 0.52} \\
CoTrain-GRACE & \cellcolor{tan}{65.15 $\pm$ 0.46} & {62.12 $\pm$ 0.33} & \cellcolor{tan}{64.13 $\pm$ 0.58} & \cellcolor{tan}{56.30 $\pm$ 0.34} & \cellcolor{brown!20}{56.97 $\pm$ 0.57} \\
% Retrieval-DGI & {61.71 $\pm$ 0.63} & {59.75 $\pm$ 0.73} & {62.26 $\pm$ 0.92} & {52.89 $\pm$ 0.78} & {56.38 $\pm$ 0.45} \\
% Retrieval-GRACE & {62.59 $\pm$ 1.19} & {60.33 $\pm$ 0.63} & {62.43 $\pm$ 1.49} & {54.90 $\pm$ 0.57} & {55.67 $\pm$ 0.73} \\
Retrieval-DGI
& {59.57 $\pm$ 0.98}
& {58.34 $\pm$ 1.03}
& {60.95 $\pm$ 0.83}
& {52.64 $\pm$ 0.98}
& {55.07 $\pm$ 0.63}
\\

Retrieval-GRACE
& {59.37 $\pm$ 0.68}
& {57.68 $\pm$ 0.75}
& {60.16 $\pm$ 0.76}
& {52.55 $\pm$ 0.77}
& {55.21 $\pm$ 0.79}
\\
\bottomrule
\end{tabular}
\end{table*}

% \begin{table}[t]
% \centering
% \small
% \setlength{\tabcolsep}{2pt}
% \caption{Training-node KNN statistics before feature smoothing on OGB-Arxiv.}
% \label{tab:arxiv-knn-before-smoothing}
% \begin{tabular}{l|cccc|cccc}
% \toprule
% \textbf{Space}
% & \multicolumn{4}{c|}{$K=30$}
% & \multicolumn{4}{c}{$K=100$} \\
% \cline{2-9}
% & $U$ & $P$ & Top-1 & Top-3
% & $U$ & $P$ & Top-1 & Top-3 \\
% \midrule
% Supervised.
% & 4.77 & 56.20\% & 67.53\% & 88.64\%
% & 8.36 & 52.16\% & 67.54\% & 88.81\% \\

% DGI-SSL
% & 10.21 & 28.75\% & 43.92\% & 65.07\%
% & 18.11 & 26.10\% & 42.54\% & 65.59\% \\

% GRACE-SSL
% & 10.19 & 26.58\% & 35.16\% & 56.26\%
% & 17.41 & 25.22\% & 35.52\% & 57.15\% \\
% \bottomrule
% \end{tabular}
% \end{table}

\begin{table}[t]
\centering
\small
\setlength{\tabcolsep}{2pt}
\caption{Neighbor-based retrieval results across supervised and SSL representation spaces on the validation set of \texttt{OGB-Arxiv}.}
\label{tab:arxiv-retrieval-val-knn}
\begin{tabular}{l|cccc|cccc}
\toprule
\textbf{Space}
& \multicolumn{4}{c|}{$K=30$}
& \multicolumn{4}{c}{$K=100$} \\
\cline{2-9} & $U$ & $P$ & Top-1 & Top-3
& $U$ & $P$ & Top-1 & Top-3 \\
\midrule
Supervised
& 5.42 & 43.06\% & 47.16\% & 72.83\%
& 9.34 & 40.58\% & 47.19\% & 73.19\% \\

DGI-SSL
& 10.89 & 23.31\% & 35.14\% & 56.22\%
& 19.22 & 21.24\% & 34.88\% & 57.04\% \\

GRACE-SSL
& 10.91 & 21.52\% & 28.97\% & 49.59\%
& 18.64 & 20.52\% & 29.43\% & 50.60\% \\
\bottomrule
\end{tabular}
\end{table}

\begin{table}[t]
\centering
\normalsize
\setlength{\tabcolsep}{6pt}
\caption{Training-node label purity before and after feature smoothing on \texttt{OGB-Arxiv} with a GAT backbone.}
\label{tab:arxiv-smoothing-purity}
\begin{tabular}{lccc}
\toprule
\textbf{Branch} & Before &After & $\Delta$\\
\midrule
GRACE SSL
& 26.78\% & 33.48\% & +6.70 pp \\
Supervised
& 55.46\% & 60.15\% & +4.69 pp \\
\bottomrule
\end{tabular}
\end{table}

\begin{figure}[t]
    \centering

    \begin{subfigure}[t]{0.32\linewidth}
        \centering
        \includegraphics[
            width=\linewidth,
            trim=0pt 0pt 520pt 0pt,
            clip
        ]{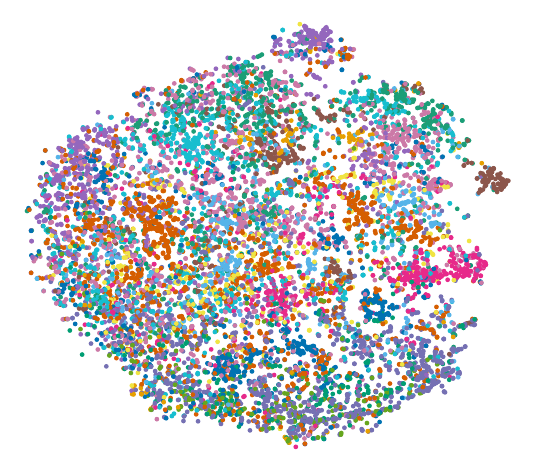}
        \caption{Supervised}
    \end{subfigure}
    \hfill
    \begin{subfigure}[t]{0.32\linewidth}
        \centering
        \includegraphics[
            width=\linewidth,
            trim=260pt 0pt 260pt 0pt,
            clip
        ]{figs/dgi_15classes.pdf}
        \caption{Self-Supervised}
    \end{subfigure}
    \hfill
    \begin{subfigure}[t]{0.32\linewidth}
        \centering
        \includegraphics[
            width=\linewidth,
            trim=520pt 0pt 0pt 0pt,
            clip
        ]{figs/dgi_15classes.pdf}
        \caption{Co-Train (DGI)}
    \end{subfigure}
    \captionsetup{name={Figure}}
    \caption{t-SNE visualization of supervised, self-supervised (SSL), and
Co-Train test-node representations on the 2018--2020 temporal split of
\texttt{OGB-Arxiv} with DGI.}
    \label{fig:tsne-dgi-appendix}
\end{figure}

\section{Additional Analysis}
\label{sec:additional_results}
\subsection{Visualization of Learned Representations.}
\label{sec:app-tsne-dgi}
Figure~\ref{fig:tsne-dgi-appendix} visualizes the supervised, self-supervised,
and Co-Train representations on \texttt{OGB-Arxiv} using t-SNE, with
DGI as the SSL objective. The corresponding quantitative analysis in
the main paper shows that CoTrain-DGI achieves the highest representation
separation score, indicating more compact intra-class representations
and greater inter-class separation. Consistent with this observation,
the t-SNE visualization shows clearer class structure after integrating
the supervised and SSL representations. Together with the quantitative
results in the main paper, this provides additional evidence that the
two representation spaces contain complementary information and that
Co-Train can integrate them into a more discriminative representation.

\subsection{Neighbor-Based Retrieval Results on \texttt{OGB-Arxiv}.}
\label{sec:app-centroid-retrieval}
We analyze neighbor-based retrieval on \texttt{OGB-Arxiv} to explain
our use of class-centroid retrieval in this multiclass setting.
We use GAT as the supervised backbone and GraphSAGE as the SSL encoder.
Table~\ref{tab:arxiv-retrieval-val-knn} reports validation results for
the supervised, DGI, and GRACE representation spaces with $K=30$ and
$K=100$.

For each query node $x$, let $\mathcal{N}_K(x)$ denote its $K$
retrieved training neighbors, where $j$ indexes a retrieved neighbor
and $y_x$ and $y_j$ denote the labels of the query node and neighbor
$j$, respectively. We characterize the retrieved neighborhood using
class diversity and label purity:
\[
U_K(x)
=
\left|\{y_j:j\in\mathcal{N}_K(x)\}\right|,
\qquad
P_K(x)
=
\frac{1}{K}
\sum_{j\in\mathcal{N}_K(x)}
\mathbb{I}[y_j=y_x].
\]
Here, $U_K(x)$ is the number of unique classes among the retrieved
neighbors, while $P_K(x)$ is the fraction of neighbors sharing the
ground-truth label of the query node. We report both quantities averaged
over all validation nodes.

We further evaluate the class-level prediction induced by the retrieved
neighbors using the same similarity-weighted aggregation as in
Dual-Space Retrieval. For representation space
$b\in\{\mathrm{sup},\mathrm{ssl}\}$, the class-score vector is
\[
\boldsymbol{\eta}_b(x)
=
\sum_{j\in\mathcal{N}_K(x,\mathcal{M}_b)}
w_{xj}^{b}\mathbf{y}_j,
\]
where $w_{xj}^{b}$ is the normalized similarity weight and
$\mathbf{y}_j$ is the one-hot label of neighbor $j$. Let
$\mathrm{Top}_m(\boldsymbol{\eta}_b(x))$ denote the $m$ classes with
the highest scores in $\boldsymbol{\eta}_b(x)$. Top-1 and Top-3 report
the percentage of validation nodes satisfying
\[
y_x \in \mathrm{Top}_1(\boldsymbol{\eta}_b(x))
\quad\text{and}\quad
y_x \in \mathrm{Top}_3(\boldsymbol{\eta}_b(x)),
\]
respectively.

The results show that neighbor-based retrieval can be highly local in
this 40-class setting, particularly in the SSL representation spaces.
With $K=30$, the DGI and GRACE neighborhoods contain approximately 11
unique classes on average, with label purity of only 23.31\% and
21.52\%, respectively. The ground-truth class appears in Top-1 for
35.14\% and 28.97\% of validation nodes, and in Top-3 for 56.22\%
and 49.59\%. Increasing $K$ to 100 expands the neighborhoods to
approximately 19 classes, but purity remains low and Top-1 performance
changes little.

These results suggest that neighbor-based retrieval can be overly local
in this multiclass setting, where the retrieved neighborhood covers only
a limited subset of classes and may not provide sufficiently
representative class-level signals. This issue is particularly evident
in the SSL spaces, where low neighborhood label purity is accompanied
by low Top-1 accuracy, indicating that the ground-truth class is often
not well represented among the retrieved neighbors. As a result,
similarity-weighted voting is restricted to the classes present in the
local neighborhood and cannot effectively compare against classes
outside it. We therefore use class-centroid retrieval on
\texttt{OGB-Arxiv}, which explicitly represents every class and enables
prediction over the full class space.

\subsection{Feature Smoothing for Centroid-Based Retrieval on
\texttt{OGB-Arxiv}.}
\label{sec:app-feature-smoothing}
After adopting class-centroid retrieval to provide explicit
representatives for all classes, we examine the quality of the
underlying representation space. In centroid-based retrieval, prediction
depends on how distinctly a node can be associated with different class
representatives. On \texttt{OGB-Arxiv}, we observe relatively low
retrieval confidence, measured by
$C_b(x)=\max_c[\boldsymbol{\eta}_b(x)]_c$, suggesting that the class-level
retrieval signal can remain ambiguous in this multiclass setting. Thus, to improve the local consistency of the representation space, we consider feature smoothing before constructing the class centroids.

Table~\ref{tab:arxiv-smoothing-purity} analyzes this effect on the
training nodes, using GAT as the supervised backbone and GraphSAGE as
the SSL encoder. We compare neighborhood label purity before and after
feature smoothing in both representation spaces. Feature smoothing
increases label purity, indicating that neighboring nodes become more
locally consistent after smoothing. This provides a more coherent
representation space for constructing class centroids and helps produce
more distinct class-level retrieval signals. Consistent with this
analysis, the main results show substantial improvements from the
feature-smoothed variant, \textit{Retrieval*}, over the corresponding
centroid-based Retrieval results.

\subsection{Efficiency and Memory Usage Comparison.}
\label{sec:app-run}
Table~\ref{tab:runtime_memory} compares the computational efficiency of
all methods on \texttt{OGB-Arxiv}. We use GraphSAGE as both the
supervised backbone and the SSL encoder for our methods, and GraphSAGE
as the backbone for all compared methods to ensure a consistent
comparison. All experiments are conducted on the same machine, and we
report the total training and inference runtime of each method under the
hyperparameter configuration that achieves its reported performance.

Co-Train is the most efficient method in our comparison, requiring only
43.75 seconds, compared with 70.83 seconds for ERM, 50.29 seconds for
LiSA, 206.19 seconds for MARIO, and more than two hours for EERM. This
efficiency is particularly notable given its strong OOD performance,
showing that incorporating complementary SSL representations through
Co-Train does not introduce substantial computational overhead.
Dual-Space Retrieval is also lightweight, requiring 69.16 seconds,
while its feature-smoothed variant requires 96.52 seconds. Both remain
substantially faster than more computationally intensive OOD methods
such as MARIO and EERM.

The retrieval runtime on \texttt{OGB-Arxiv} is relatively low because
we use class-centroid retrieval, where only a small number of class
representatives are involved in retrieval. For datasets using
neighbor-based retrieval, the retrieval cost increases with the number
of candidate neighbors and retrieved neighbors $K$. This additional
cost is confined to the retrieval stage and does not increase model
training time. All evaluated methods fit within GPU memory, indicating
that our proposed frameworks introduce no prohibitive memory overhead.
% \section{Computation Efficiency and Memory Usage}
% \label{sec:appendix_memory}
% We evaluate the computational efficiency and memory usage of different methods using the \texttt{OGB-Arxiv} dataset as a representative example. For efficiency, we report the total runtime for a single training run. The comparison is shown in Table \ref{tab:runtime_memory}. Among all methods, \textbf{Co-train} is the most efficient, completing in approximately 36 seconds. This reflects its design, which integrates supervised and unsupervised training with minimal overhead. \textbf{LiSA} exhibits comparable runtime to \textbf{ERM}, despite incorporating additional steps such as subgraph sampling and alignment, due to their relatively lightweight nature. The \textbf{Centroid-based Retrieval Voting} approach incurs higher runtime, as it trains two separate models and involves distance-based retrieval, which is computationally more intensive. However, its runtime remains well within practical limits and is significantly lower than that of \textbf{EERM}. EERM relies on reinforcement learning to generate augmented graph environments, which introduces substantial computational and memory overhead, often exceeding the capacity of a 16GB GPU. Overall, while Retrieval Voting is slower than ERM-style baselines, it remains competitive in efficiency and considerably more tractable than reinforcement-learning-based approaches like EERM.

\begin{table}[t]
\centering
\small
\caption{Runtime and memory usage across methods on the
\texttt{OGB-Arxiv} dataset using a backbone.}
\label{tab:runtime_memory}
\resizebox{\columnwidth}{!}{%
\begin{tabular}{l|cc}
\toprule
\textbf{Method}
& Total Runtime (s)
& Fits in GPU Memory\\
\midrule
ERM                         & 70.83      & Yes \\
EERM                        & $>2$ hours & Yes \\
LiSA                        & 50.29      & Yes \\
MARIO                       & 206.19     & Yes \\
Co-Train                    & 43.75      & Yes \\
Dual-Space Retrieval                   & 69.16      & Yes \\
Dual-Space Retrieval\textsuperscript{*}& 96.52      & Yes \\
\bottomrule
\end{tabular}%
}
\end{table}
\section{Sensitivity and Robustness Analysis}
\label{sec:app-sensitivity_analysis}

\subsection{Sensitivity to the SSL Fusion Weight for Dual-Space Retrieval.}
\label{sec:app-sensitive_fusion_weight}
Table~\ref{tab:arxiv-epsilon-ablation} evaluates the sensitivity to
the SSL fusion weight $\alpha_{\mathrm{ssl}}$ in Dual-Space Retrieval
on \texttt{OGB-Arxiv}, using GAT as the supervised backbone and GRACE
as the SSL encoder. As described in the main paper, the two retrieval
branches exhibit large prediction disagreement ($\delta>0.5$) on this
dataset. We therefore use a small fixed $\alpha_{\mathrm{ssl}}$ and
vary its value over $\{0,0.01,0.05,0.1\}$. Performance remains similar
across all three temporal splits; for example, accuracy on 2018--2020
ranges from 44.92\% to 45.25\%. This shows that the results are not
sensitive to the exact choice of a small SSL fusion weight.  We therefore set $\alpha_{\mathrm{ssl}}=0.01$ globally for the
\texttt{OGB-Arxiv} experiments.
\begin{table}[t]
\centering
\normalsize
\setlength{\tabcolsep}{3pt}
\caption{Test accuracy (\%) under different SSL fusion coefficients $\alpha_{ssl}$ on \texttt{OGB-Arxiv} using Retrieval-GRACE with a GAT backbone.}
\label{tab:arxiv-epsilon-ablation}

\begin{tabular}{c|ccc}
\toprule
\textbf{$\alpha_{ssl}$} & 2014--2016 & 2016--2018 & 2018--2020 \\
\midrule
0
& 47.75 $\pm$ 0.22
& 47.07 $\pm$ 0.22
& 44.92 $\pm$ 0.30 \\
0.01
& 47.71 $\pm$ 0.21
& 47.02 $\pm$ 0.21
& 45.04 $\pm$ 0.18 \\
0.05
& 47.96 $\pm$ 0.23
& 47.25 $\pm$ 0.31
& 45.18 $\pm$ 0.31 \\
0.1
& 47.91 $\pm$ 0.25
& 47.33 $\pm$ 0.29
& 45.25 $\pm$ 0.23 \\
\bottomrule
\end{tabular}
\end{table}

\begin{table}[t]
\centering
\normalsize
\setlength{\tabcolsep}{6pt}
\caption{Test accuracy (\%) with different numbers of retrieved centroids $K$ on \texttt{OGB-Arxiv} using Retrieval-GRACE with a GAT backbone.} 
\label{tab:arxiv-centroid-k-ablation}

\begin{tabular}{c|ccc}
\toprule
\textbf{$K$} & 2014--2016 & 2016--2018 & 2018--2020 \\
\midrule
10
& {48.04 $\pm$ 0.13}
& {47.32 $\pm$ 0.11}
& {45.17 $\pm$ 0.20} \\

20
& {47.83 $\pm$ 0.11}
& {47.21 $\pm$ 0.13}
& {45.07 $\pm$ 0.03} \\

30
& {47.87 $\pm$ 0.08}
& {47.19 $\pm$ 0.23}
& {45.13 $\pm$ 0.04} \\

40
& {47.71 $\pm$ 0.21}
& {47.16 $\pm$ 0.32}
& {45.33 $\pm$ 0.46} \\

\bottomrule
\end{tabular}

\end{table}

\subsection{Sensitivity to the Retrieval Size $K$.}
\label{sec:app-sensitive_k}
We further examine the sensitivity of Dual-Space Retrieval to the
retrieval size $K$ under both centroid-based and neighbor-based
retrieval.

Table~\ref{tab:arxiv-centroid-k-ablation} reports results on
\texttt{OGB-Arxiv} using GAT as the supervised backbone and GRACE as
the SSL encoder. We vary $K\in\{10,20,30,40\}$ for class-centroid
retrieval and observe comparable performance across all three temporal
splits. The value used in the main experiments is selected on the
validation set. Since \texttt{OGB-Arxiv} contains 40 classes, $K=40$
corresponds to retrieving all class centroids.

Table~\ref{tab:elliptic-sage-grace-k-testwise} further evaluates $K$ for
neighbor-based retrieval on \texttt{Elliptic}, using GraphSAGE as the
supervised backbone and GRACE as the SSL encoder. We use $K=100$ based
on validation performance and compare
$K\in\{30,100,200,300,500\}$ across the nine test splits. Performance
remains stable over a wide range of $K$, with only small variations
across the test splits. Together, these results show that Dual-Space
Retrieval is not sensitive to the exact choice of $K$ under either
centroid- or neighbor-based retrieval.
\begin{table}[!t]
\centering
\small
\setlength{\tabcolsep}{3pt}
\renewcommand{\arraystretch}{1.05}
\caption{Test-wise sensitivity of Elliptic-SAGE-GRACE to the number of retrieved neighbors $K$. Each entry reports the mean F1 score (\%) and standard deviation over 10 runs.}
\label{tab:elliptic-sage-grace-k-testwise}
\begin{tabular}{l|ccccc}
\toprule
\textbf{Split}
& \textbf{$K=30$}
& \textbf{$K=100$}
& \textbf{$K=200$}
& \textbf{$K=300$}
& \textbf{$K=500$} \\
\midrule
Test1
& 92.64 $\pm$ 1.17
& 92.62 $\pm$ 0.87
& 92.10 $\pm$ 0.91
& 92.34 $\pm$ 1.09
& 92.02 $\pm$ 1.32 \\

Test2
& 84.34 $\pm$ 0.52
& 84.48 $\pm$ 0.70
& 84.29 $\pm$ 0.67
& 84.65 $\pm$ 0.77
& 84.47 $\pm$ 0.64 \\

Test3
& 78.01 $\pm$ 1.10
& 78.27 $\pm$ 0.76
& 77.92 $\pm$ 0.64
& 78.01 $\pm$ 0.75
& 78.09 $\pm$ 1.06 \\

Test4
& 74.44 $\pm$ 0.60
& 74.69 $\pm$ 0.41
& 74.41 $\pm$ 0.40
& 74.44 $\pm$ 0.45
& 74.45 $\pm$ 0.35 \\

Test5
& 72.55 $\pm$ 0.72
& 72.98 $\pm$ 0.72
& 72.74 $\pm$ 0.68
& 72.64 $\pm$ 0.70
& 72.57 $\pm$ 0.99 \\

Test6
& 80.33 $\pm$ 0.97
& 80.44 $\pm$ 1.05
& 80.89 $\pm$ 0.73
& 80.48 $\pm$ 1.08
& 80.54 $\pm$ 1.26 \\

Test7
& 78.14 $\pm$ 1.34
& 77.83 $\pm$ 1.32
& 78.24 $\pm$ 1.52
& 78.09 $\pm$ 1.48
& 77.84 $\pm$ 1.36 \\

Test8
& 66.89 $\pm$ 1.33
& 66.54 $\pm$ 1.43
& 67.34 $\pm$ 1.74
& 67.12 $\pm$ 1.73
& 67.02 $\pm$ 1.52 \\

Test9
& 50.57 $\pm$ 0.71
& 50.86 $\pm$ 1.33
& 50.44 $\pm$ 0.99
& 50.80 $\pm$ 0.85
& 50.82 $\pm$ 0.87 \\
\bottomrule
\end{tabular}
\end{table}

\subsection{Sensitivity to the SSL Encoder Architecture.}
\label{sec:app-ssl-encoder-sensitivity}
We further examine the sensitivity of Co-Train-DGI to the SSL encoder
architecture. Since the supervised and SSL encoders are separately
parameterized without parameter sharing, their backbone architectures
can be selected independently. Tables~\ref{tab:ssl-backbone-arxiv-tests} and~\ref{tab:ssl-backbone-elliptic-tests} compare GCN and GraphSAGE as the
DGI encoder on \texttt{Elliptic} and \texttt{OGB-Arxiv}, respectively.
Based on validation performance, we use GCN for \texttt{Elliptic} and
GraphSAGE for \texttt{OGB-Arxiv} in the main experiments. Dual-Space
Retrieval uses the same selected supervised and SSL backbone
configurations.

On \texttt{Elliptic}, the two SSL encoders achieve comparable performance
across the nine test splits, with each performing slightly better on
different periods. On \texttt{OGB-Arxiv}, GCN performs slightly better
on the three test periods, by only 0.41, 0.37, and 0.58 percentage
points, respectively, despite GraphSAGE being selected based on
validation performance. Overall, the small differences across both
datasets suggest that Co-Train is not highly sensitive to the choice of
SSL encoder architecture.
\begin{table}[!t]
\centering
\normalsize
\caption{Test accuracy (\%) on \texttt{OGB-Arxiv} with different SSL encoder architectures for Co-Train-DGI.}
\label{tab:ssl-backbone-arxiv-tests}

\begin{tabular}{l|cc}
\toprule
\textbf{Test split} & SSL-GCN & SSL-SAGE\\
\midrule
2014-2016 & 50.70 $\pm$ 0.28 & 50.29 $\pm$ 0.34 \\
2016-2018 & 49.41 $\pm$ 0.32 & 49.04 $\pm$ 0.48 \\
2018-2020 & 46.71 $\pm$ 0.42 & 46.13 $\pm$ 0.50 \\
\bottomrule
\end{tabular}
\end{table}

\begin{table}[t]
\centering
\normalsize
\caption{F1-score (\%) of Co-Train-DGI to the SSL encoder architecture on \texttt{Elliptic}.}
\label{tab:ssl-backbone-elliptic-tests}

\begin{tabular}{l|cc}
\toprule
\textbf{Test split} & SSL-GCN& SSL-SAGE \\
\midrule
Test1 & 94.57 $\pm$ 0.81 & 94.81 $\pm$ 0.58 \\
Test2 & 87.17 $\pm$ 0.92 & 87.41 $\pm$ 1.09 \\
Test3 & 79.50 $\pm$ 1.29 & 79.59 $\pm$ 1.16 \\
Test4 & 74.39 $\pm$ 0.83 & 74.12 $\pm$ 1.40 \\
Test5 & 79.05 $\pm$ 2.10 & 78.86 $\pm$ 2.00 \\
Test6 & 84.98 $\pm$ 2.18 & 85.38 $\pm$ 1.93 \\
Test7 & 85.93 $\pm$ 1.97 & 86.42 $\pm$ 1.69 \\
Test8 & 71.32 $\pm$ 4.70 & 73.10 $\pm$ 2.79 \\
Test9 & 50.03 $\pm$ 0.52 & 49.90 $\pm$ 0.30 \\
\bottomrule
\end{tabular}
\end{table}

\section{Additional Experimental Results}
\label{sec:additional_experiment_results}

Table~\ref{tab:app-fb100-texas} reports results on \texttt{FB-100} with
three training-domain combinations and Texas as the test domain, using
GCN as both the supervised backbone and SSL encoder.
Table~\ref{tab:twitch-gat} reports results on \texttt{Twitch} with GAT
as the supervised backbone and GCN as the SSL encoder.

On \texttt{Twitch}, Co-Train achieves the best performance on four of
the five target domains, with gains of 1.51, 0.06, 2.55, and 0.72
percentage points over the strongest existing baseline on ES, FR, PTBR,
and RU, respectively. Dual-Space Retrieval is less competitive under
the GAT backbone, suggesting that retrieval-based prediction can be
more sensitive to the learned representation space.

On \texttt{FB-100}, performance varies more noticeably across the three
training-domain combinations. GRACE achieves the best results under
JHU+CIT+AMH and BIN+DUK+PRI, with accuracies of 56.11\% and 56.08\%, respectively.
Under WUSTL+BRD+CMU, CoTrain-DGI achieves the best overall accuracy of 56.89\%,
outperforming the strongest baseline by 0.26 percentage points.

\begin{table}[!t]
\centering
\small
\setlength{\tabcolsep}{1pt}
\caption{Accuracy (\%) on the Texas of FB-100 under Different Training Graph Combinations.}
\label{tab:app-fb100-texas}
\begin{tabular}{l|ccc}
\toprule
\textbf{Method}
&\multicolumn{3}{c}{Texas}\\
\cmidrule(lr){2-4}
&{\footnotesize JHU+CIT+AMH} & {\footnotesize BIN+DUK+PRI} & {\footnotesize WUSTL+BRD+CMU}\\
\midrule
ERM & \cellcolor{tan}56.25 $\pm$ 0.01 & 52.12 $\pm$ 5.52 & 55.99 $\pm$ 0.53 \\
EERM & 52.68 $\pm$ 3.19 & 52.05 $\pm$ 5.94 & 53.83 $\pm$ 5.33 \\
LiSA & 53.01 $\pm$ 4.59 & 55.26 $\pm$ 1.48 & 54.52 $\pm$ 3.88 \\
MARIO & 50.18 $\pm$ 3.80 & \cellcolor{brown!20}55.77 $\pm$ 0.74 & 55.72 $\pm$ 1.42 \\
DGI & {46.46 $\pm$ 3.67} & {48.02 $\pm$ 1.38} & {46.19 $\pm$ 2.76} \\
GRACE & \cellcolor{brown!20}{56.11 $\pm$ 0.60} & \cellcolor{tan}{56.08 $\pm$ 0.32} & \cellcolor{brown!20}{56.63 $\pm$ 0.22} \\
CoTrain-DGI & {53.54 $\pm$ 1.92} & {53.18 $\pm$ 2.11} & \cellcolor{tan}{56.89 $\pm$ 0.40} \\
CoTrain-GRACE
& {55.08 $\pm$ 1.61}
& {53.12 $\pm$ 2.98}
& {56.60 $\pm$ 0.79} \\
% Retrieval-DGI
% & {53.36 $\pm$ 0.89}
% & {55.46 $\pm$ 0.24}
% & {56.01 $\pm$ 0.39} \\
% Retrieval-GRACE & {50.22 $\pm$ 1.19} & {50.41 $\pm$ 0.37} & {55.18 $\pm$ 0.46} \\
Retrieval-DGI
& {53.65 $\pm$ 0.85}
& {55.29 $\pm$ 0.39}
& {55.96 $\pm$ 0.35} \\

Retrieval-GRACE
& {51.21 $\pm$ 0.47}
& {52.49 $\pm$ 0.87}
& {51.80 $\pm$ 0.73} \\
\bottomrule
\end{tabular}

\end{table}

\end{document}